%% file: ACEvo_arxiv_fixed_reference_colors_compact.tex
\documentclass{article} 
\usepackage[final]{colm2026_conference}

\input{math_commands.tex} 
\usepackage{amsmath}
\usepackage{amssymb}
\usepackage{mathtools}
\usepackage{amsthm}
\usepackage{graphicx}
\usepackage{subcaption}
\usepackage{booktabs}
\usepackage{multirow}
\usepackage{array}
\usepackage{adjustbox}
\usepackage{makecell}
\usepackage{siunitx}
\usepackage{hyperref}
\usepackage{float}
\usepackage{placeins}
\usepackage{enumitem}
\usepackage{needspace}

\usepackage[table]{xcolor}
\definecolor{mistblue}{RGB}{225,235,245}

\usepackage[ruled,linesnumbered]{algorithm2e}
\DontPrintSemicolon
\SetKwInput{Input}{Input}
\SetKwInput{Output}{Output}
\SetKwComment{tcc}{// }{}
\SetKw{To}{to}
\SetKw{Init}{Init}

\usepackage{tikz}
\usetikzlibrary{plotmarks}
\usepackage{pgfplots}
\pgfplotsset{compat=1.18}
\usepgfplotslibrary{fillbetween}

\usepackage{listings}
\usepackage{tcolorbox}

\usepackage{xurl}
\usepackage{url}

\usepackage[capitalize,noabbrev]{cleveref}

\theoremstyle{plain}

\theoremstyle{definition}

\theoremstyle{remark}

\usepackage{microtype}
\usepackage{hyperref}
\usepackage{url}
\usepackage{booktabs}
\usepackage{enumitem}

\usepackage{lineno}

\definecolor{darkblue}{rgb}{0, 0, 0.5}
\hypersetup{colorlinks=true, citecolor=darkblue, linkcolor=darkblue, urlcolor=darkblue}

\title{ACEvo: Adversarial Co-Evolution of Problem Distributions and Solvers for Combinatorial Optimization}

\author{
Ruibo Duan$^{1}$, Yuxin Liu$^{1}$, Haoran Ye$^{2}$,
Xinyao Dong$^{1}$, Zhiqiang Xu$^{3}$,
Chenglin Fan$^{1,\dagger}$ \\
$^{1}$Department of Computer Science and Engineering, Seoul National University \\
$^{2}$School of Intelligence Science and Technology, Peking University \\
$^{3}$Mohamed bin Zayed University of Artificial Intelligence (MBZUAI)
}

\begin{document}

\ifcolmsubmission
\linenumbers
\fi

\maketitle

\begingroup
\renewcommand{\thefootnote}{\fnsymbol{footnote}}
\footnotetext[2]{Corresponding author.}
\endgroup
\begin{abstract}
Large language models (LLMs) are increasingly used to synthesize heuristic programs, yet most existing pipelines optimize solvers against fixed benchmark distributions. This static setup can obscure solver weaknesses and limit understanding of how LLM-designed algorithms adapt under distribution shift. We present Adversarial Co-Evolution(\texttt{ACEvo}), a closed-loop framework in which LLMs iteratively co-evolve two types of executable programs: heuristic solvers and problem generators. The generator proposes increasingly challenging instances, while the solver is refined to improve performance on the evolving distribution, forming an automated adversarial curriculum for program design and evaluation. We instantiate \texttt{ACEvo} on routing problems, including TSP, OP, and CVRP. Across these domains, the framework produces instance distributions that consistently induce larger optimality gaps than standard benchmarks and yields solver programs that outperform those obtained from static-training baselines under distribution shift. Beyond final performance, \texttt{ACEvo} provides a testbed for studying LLM-based algorithm design under evolving distributions, including how reflective mutation, adversarial feedback, and co-adaptation shape the evolution of both generators and solvers. These results suggest that closed-loop co-evolution is a promising paradigm for using language models not only to generate algorithms, but also to construct adaptive evaluation environments.
\end{abstract}

\section{Introduction}

Combinatorial optimization (CO) problems are central to Web-scale systems, underpinning tasks such as large-scale scheduling, content delivery, traffic routing, and personalized recommendation. The exponential growth of solution spaces and the heterogeneity of Web environments make the design of robust and scalable solvers a persistent challenge~\cite{GareyJohnson1979,GendreauPotvin2005}. Traditional approaches rely on handcrafted heuristics, which are costly to design and difficult to adapt across diverse Web applications. Recently, large language models (LLMs) have demonstrated strong capabilities in code generation and complex reasoning, opening a promising direction for automating solver design tailored to Web-scale optimization~\cite{ChenCodex21,Li2022AlphaCode}.

Existing studies typically rely on fixed datasets or uniform instance generation protocols, resulting in limited diversity across problem settings. Moreover, most approaches focus on incremental enhancements to existing solvers rather than introducing fundamentally novel frameworks~\cite{YeReEvo24,Bomer2025LLMHeuristics,SartoriBlum2025LLMImprove,mu2025planning}. This static evaluation paradigm restricts our ability to assess solver generalization and robustness. As a result, two critical challenges remain underexplored: (1) how to generate increasingly challenging problem instances, and (2) how to design solvers that can co-evolve alongside rising problem complexity. Addressing these challenges requires a fully automated, mutually reinforcing framework that can simultaneously synthesize harder instances and adapt solver strategies in response.

To address these challenges, we introduce \texttt{ACEvo} (Adversarial Co-Evolution of Problem Distributions and Solvers), a novel framework that leverages large language models to co-evolve optimization problem instances and their corresponding heuristic solvers. \texttt{ACEvo} iteratively mutates instance-generation programs to produce increasingly complex and difficult problems, while concurrently synthesizing more adaptive solver code. This adversarial co-evolution loop enables the solver and problem to continuously challenge and improve each other.

Experimental results show that \texttt{ACEvo} can generate problem instances that are significantly more challenging than existing evaluation instance distributions, while simultaneously synthesizing heuristic solvers with strong performance and generalization capabilities—all achieved without manual intervention. This framework paves a new path for LLM-driven instance distribution generation and solver co-evolution in combinatorial optimization, enabling a dynamic evaluation paradigm that more faithfully assesses solver robustness and adaptability under increasing problem complexity.
The proposed framework admits a more general formulation, as discussed in Appendix~\ref{appendix:extended_applications}.

We summarize our key contributions as follows:

\begin{itemize}[leftmargin=*, nosep]
\item We introduce \textbf{ACEvo}, a novel framework that enables fully automated co-evolution of optimization instances and solver algorithms via large language models.
\item We develop a mutation-based adversarial mechanism that adaptively modifies instance generation procedures to yield increasingly challenging optimization problems.

\item We propose a structured generation process that leverages large language model to automatically synthesize adaptive, high-quality heuristic algorithms without human intervention.
\item We empirically validate \textbf{ACEvo} across multiple combinatorial optimization tasks, demonstrating its broad effectiveness and robustness on classical TSPLIB~\cite{Reinelt91} instances, and its capability to generate more challenging problem instances and synthesize solvers that achieve state-of-the-art performance.
\end{itemize}

\vspace{-0.5em}

\section{Related Work}

In this section, we summarize related works, including classical heuristics, instance distribution construction, adversarial optimization, and LLM-based methods.
\vspace{-1em}
\paragraph{Classical and Heuristic Methods for Combinatorial Optimization.}

Classical approaches to combinatorial optimization can be broadly categorized into exact algorithms and heuristic methods. Exact algorithms such as branch-and-bound~\cite{PadbergR91} guarantee optimal solutions but suffer from poor scalability as problem size increases. To address this, heuristic approaches like greedy algorithms and local search are often employed to provide efficient, though suboptimal, solutions. Metaheuristics—such as simulated annealing~\cite{SkiscimG83} and ant colony optimization~\cite{DorigoG97}—offer improved scalability on TSP and VRP~\cite{BlumR03,LiuLu2023}, while graph-based learning supports structured planning and correspondence tasks~\cite{ma2023planning,zhang2024fastmac}.
\vspace{-1em}

\paragraph{Instance Distribution Design and Difficulty Modeling.}

Beyond designing effective solvers, constructing instance distributions with high discriminatory power is critical for evaluating and comparing algorithm performance~\cite{SmithMilesH11}. While the widely-used benchmark TSPLIB~\cite{Reinelt91} has been a standard in the literature~\cite{SunY23,LuoLiu23,LiGWY23}, it does not support instance difficulty adjustment based on solver capabilities. Other datasets generated by fixed-range random sampling often yield uniform or easy instances that lack diversity and fail to challenge solvers across a range of difficulty levels~\cite{SmithMilesH11}. On such datasets, solvers frequently converge to similar optima, making it difficult to assess their relative strengths~\cite{LiuICML24,Yao2025MOEHeuristicLLM,RomeraParedes2024MathDiscoveryLLM}. Data-centric studies also emphasize long-tailed difficulty and sample value~\cite{xiao2026curiosity,xiao2026points}, while larger instances involve complex structural patterns~\cite{applegate2011traveling}.
In contrast, our proposed \texttt{ACEvo} framework actively generates challenging problem instances that significantly amplify performance differences across solvers.

\vspace{-0.5em}
\paragraph{Adversarial Training in Combinatorial Optimization.}

Adversarial training, originally developed in the context of image generation via Generative Adversarial Networks (GANs)~\cite{GoodfellowGAN2014}, has shown promise in synthesizing high-quality, diverse data~\cite{ZhuCycleGAN17,LedigSRGAN2017}. Inspired by this success, recent research has applied adversarial ideas to combinatorial optimization~\cite{XinGANCO2022,WangGIRL23,HeGANMO2021,GuoMOGAN2020}, primarily to improve solution quality or enhance solver generalization. These approaches typically focus on optimizing solutions via adversarial learning objectives. In contrast, our work introduces a co-evolutionary adversarial framework that simultaneously evolves both instance generators and solvers.

\vspace{-0.5em}
\paragraph{Large Language Models for Optimization.}

Recent advances have shown the potential of large language models (LLMs) in tackling a variety of optimization and structured prediction tasks by leveraging their strong generalization and program synthesis capabilities~\cite{ZhaoVizOpt25,JiangDSM24,WuASLLM23,ZhaoTaskPlanning23,iklassov2024selfguidingexplorationcombinatorialproblems,zheng2024large}. Common strategies include few-shot prompting~\cite{BrownGPT3}, code completion~\cite{ChenCodex21}, and feedback-guided iterative refinement~\cite{LiuICML24,YeReEvo24,ZelikmanSTaR22,lin2026cec}. Within combinatorial optimization, LLMs are primarily used as solvers~\cite{YangLLMOpt24,LiuLLMEvoOpt24,ZhaoTrustLLM25,TangPromptOpt25}, generating and refining candidate solutions through code synthesis. They are often integrated with evolutionary techniques such as genetic algorithms to iteratively improve solution quality~\cite{liu2023algorithm,LiuICML24,YeReEvo24}, demonstrating growing potential as hybrid learning and search agents.

\vspace{-0.5em}
\section{Method}
\vspace{-0.5em}

In this section, we present \texttt{ACEvo}, our evolutionary adversarial framework that jointly generates hard combinatorial optimization instances and synthesizes adaptive heuristic solvers. The method consists of four main components: (i) an LLM-based hard instance generator that mutates procedural programs to create increasingly challenging problems, (ii) an LLM-based heuristic generator that produces solver algorithms guided by symbolic prompts, (iii) an interaction protocol that formulates their competition as a symbolic adversarial game, and (iv) a co-evolutionary mechanism that couples solver improvement with instance difficulty escalation. The following subsections provide detailed explanations of these components and illustrate how \texttt{ACEvo} enables automated, feedback-driven co-evolution of instances and solvers.
\vspace{-0.5em}

\subsection{LLM-Based Generators for Instances and Heuristics}
\texttt{ACEvo} represents both instance generation and heuristic synthesis as LLM-generated executable programs. The following paragraphs formalize these two generators and the performance signals used to guide their evolution.

\vspace{-0.5em}

\paragraph{LLM-Based Hard Instance Generator.}
Let \( \mathcal{G} \) denote the space of instance generators \( g: \mathcal{Z} \rightarrow \mathcal{I} \), where \( \mathcal{Z} \) is a latent seed space and \( \mathcal{I} \) represents the space of problem instances (e.g., TSP instances). Each generator \( g \in \mathcal{G} \) is an executable program synthesized by a language model \( \mathcal{L}_{\text{inst}} \), conditioned on a natural language prompt \( \mathcal{P}_{\text{inst}} \). Formally,
\begin{equation}
    g \sim \mathcal{L}_{\text{inst}}(\mathcal{P}_{\text{inst}}), \quad I = g(z), \; z \sim \mathcal{Z}.
\end{equation}
The resulting generator defines a programmatic distribution over instances through algorithmic logic, spatial sampling patterns, and geometric transformations.

To quantify instance difficulty in a normalized and scale-invariant manner, we measure the \emph{relative optimality gap} with respect to a solver \( h \):
\begin{equation}
    \text{Gap}(I; h) = \frac{\mathbb{E}_{i=1}^{n} \left[ f(h(I_i)) \right]}{\mathbb{E}_{i=1}^{n} \left[ f^*(I_i) \right]} - 1,
\end{equation}
where \( \{I_1, \dots, I_n\} \) are instances sampled from the generator \( g \), \( f(h(I_i)) \) denotes the heuristic solution cost, and \( f^*(I_i) \) is the optimal or best-known reference cost. Larger gap values indicate greater instance hardness relative to solver \( h \).

\paragraph{LLM-Based Heuristic Generator.}
Let \( \mathcal{H} \) denote the space of heuristic solvers \( h: \mathcal{I} \rightarrow \mathcal{S} \), where \( \mathcal{S} \) is the solution space (e.g., feasible tours). Each heuristic \( h \in \mathcal{H} \) is synthesized as an executable decision procedure by a language model \( \mathcal{L}_{\text{heur}} \), conditioned on a natural language prompt \( \mathcal{P}_{\text{heur}} \):
\begin{equation}
    h \sim \mathcal{L}_{\text{heur}}(\mathcal{P}_{\text{heur}}).
\end{equation}
The prompt encodes algorithmic priors such as greedy strategies, local refinement rules, or metaheuristic principles, together with task-specific constraints.

Once generated, the heuristic is evaluated behaviorally by applying it to instances \( I \in \mathcal{I} \) and measuring its objective value \( f(h(I)) \). Importantly, heuristics are treated as black-box symbolic programs: no structural assumptions or manual interventions are imposed. Selection and refinement are driven solely by performance feedback, enabling the emergence of non-trivial and potentially unconventional algorithmic behaviors.

\subsection{Co-Evolution Through Symbolic Adversarial Interaction}

We formalize the learning process as a symbolic co-evolutionary game between an instance constructor \( g \in \mathcal{G} \) and a heuristic solver \( h \in \mathcal{H} \). This interaction is guided by a minimax formulation:

\begin{equation}
    \min_{h \sim \mathcal{L}_{\text{heur}}} \; \max_{g \sim \mathcal{L}_{\text{inst}}} \; \mathbb{E}_{z \sim \mathcal{Z}} \left[ \mathcal{H}(g(z); h) \right],
\end{equation}

where \( \mathcal{H}(g(z); h) \) denotes the relative hardness of the instance \( I = g(z) \) with respect to solver \( h \). This framework captures the adversarial dynamics between instance generation and solution strategies: the generator evolves to reveal vulnerabilities in current heuristics, while the solver adapts to counter increasingly complex instance structures.
\vspace{-1em}
\paragraph{Interaction Protocol.} The co-evolution proceeds through iterative interaction in the algorithmic function space:

\begin{enumerate}[leftmargin=1.7em, itemsep=0.3em, topsep=0.2em, parsep=0pt, partopsep=0pt]
    \item \textbf{Heuristic synthesis}: Sample a solver \( h \sim \mathcal{L}_{\text{heur}}(\mathcal{P}_{\text{heur}}) \) from the heuristic model via symbolic prompt programming;
    \item \textbf{Hard instance constructor synthesis}: Sample an instance generator \( g \sim \mathcal{L}_{\text{inst}}(\mathcal{P}_{\text{inst}}) \), which defines a distribution over problem instances via structured algorithmic composition under adversarial evolutionary pressure throughout co-evolution;

    \item \textbf{Evaluation}: Apply \( h \) to a batch of instances \( \{ I_i = g(z_i) \}_{i=1}^{n} \), compute the relative hardness \( \mathcal{H}(I_i; h) \), and identify instances exhibiting maximal challenge;
\end{enumerate}

This adversarial loop induces an emergent curriculum over the symbolic policy space: instance constructors synthesize increasingly diverse and non-trivial spatial structures, while solvers develop greater resilience and generalization capabilities. Notably, this process operates entirely without gradient-based optimization, relying instead on behavioral evaluation over executable algorithmic trajectories.

\subsection{Architecture and Evolutionary Process of \texttt{ACEvo}}

\label{gen_inst}

\begin{figure*}[t] 
  \centering
  \includegraphics[width=\textwidth]{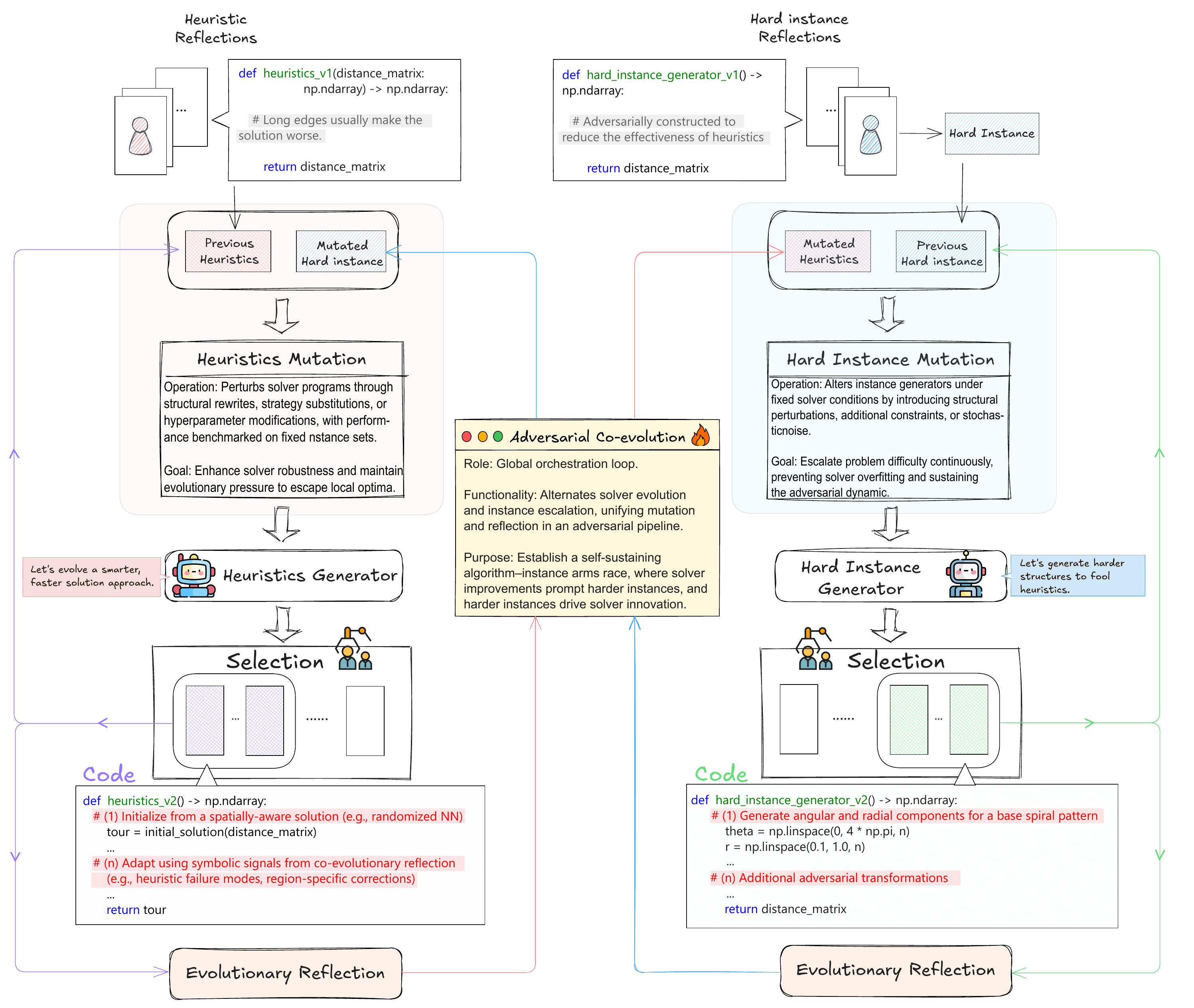}
  \caption{Illustration of the \texttt{ACEvo} framework and generated instance examples.}
  \label{fig:framework}
\end{figure*}

We propose \texttt{ACEvo}, a dual-loop adversarial co-evolution framework that simultaneously evolves hard combinatorial optimization instances and their corresponding solver heuristics. As illustrated in Figure~\ref{fig:framework}, the framework integrates two tightly coupled processes instance evolution and solver evolution both orchestrated by large language models (LLMs) and driven by adversarial feedback. In each iteration, solvers are generated, evaluated, and refined through evolutionary reflection, while instance generators are mutated in parallel to exploit solver weaknesses and escalate difficulty (see Appendix~\ref{app:acevo-pseudocode} for full pseudocode). This dynamic interaction forms a continual adversarial loop, where solvers progressively acquire stronger strategies and instances evolve into increasingly complex structures. Figure~\ref{fig:framework} provides an overview of this co-evolutionary cycle, highlighting how mutation, reflection, and selection jointly sustain the arms race between solvers and instances. A detailed analysis of the evolutionary process, including representative generator and heuristic variants across iterations, is provided in Appendix~\ref{subsec:evolution-process}.

\textbf{Heuristics generator.} The heuristics generator in \texttt{ACEvo} is an LLM-based component that synthesizes candidate solver programs as executable code fragments. These solvers are conditioned on problem descriptions and prior performance 
reflections, and are sampled from an open-ended algorithmic space without relying on predefined templates. This allows for the emergence of novel search strategies and scoring mechanisms throughout evolution.
\FloatBarrier
\textbf{Hard instance generator.} The hard instance generator plays a symmetric role in \texttt{ACEvo} by producing instance-generation programs. Each generator defines a procedural method to construct challenging combinatorial problem instances, such as spatially clustered or structurally irregular TSP variants. These generators are optimized to exploit solver weaknesses, gradually shaping a landscape of increasing difficulty. This, in turn, causes solvers to exhibit significantly different performance on the generated hard instances, providing valuable guidance for subsequent optimization by highlighting robustness gaps and informing targeted improvement strategies.

\textbf{Selection.}  
\texttt{ACEvo} employs a selection mechanism to determine which solvers and instance generators proceed to the next evolutionary round.  
For solver evolution, candidates with superior performance on current hard instances are preferentially retained, while a small fraction of diverse solvers is preserved to maintain exploration.  
For instance evolution, generators that induce higher relative hardness are selected, subject to structural diversity constraints.  
This bidirectional selection sustains adversarial pressure and prevents premature convergence, enabling a continual escalation between solvers and instances.

\textbf{Heuristics mutation.} Given a fixed hard instance, \texttt{ACEvo} selects a previously generated solver and mutates it via the heuristics generator. Mutations may include algorithmic rewrites, strategy switching, or hyperparameter changes. The resulting solver is then evaluated on the current instance and compared to its predecessor to determine its relative effectiveness.

\textbf{Hard instance mutation.} Conversely, given a fixed solver, \texttt{ACEvo} mutates an existing instance generator using the hard instance generator. The objective is to synthesize instances that are more difficult for the solver, thereby increasing selection pressure and encouraging solver robustness.

\textbf{Evolutionary reflection.} \texttt{ACEvo}  adopts evolutionary reflection, a symbolic diagnostic mechanism wherein LLMs evaluate behavioral shifts between original and mutated programs to generate high-level feedback—such as failure modes, inefficiencies, or refinement cues. These insights are embedded into future prompts, enabling implicit memory and long-term adaptation. Through this mechanism, \texttt{ACEvo} selectively produces mutated heuristics or hard instances based on task-specific objectives, allowing the co-evolutionary process to dynamically align with either solver optimization or instance difficulty escalation.

\textbf{Adversarial co-evolution process.} The two evolutionary paths in \texttt{ACEvo} are tightly coupled through adversarial interaction. As solvers improve, instance generators are pressured to synthesize harder instances; in turn, more difficult instances expose limitations in solver logic. This continuous feedback loop creates an arms race between the two components, driving the emergence of increasingly generalizable solvers and increasingly sophisticated instance distributions.

\section{Experiments}

We conduct experiments to evaluate the effectiveness of \texttt{ACEvo} across combinatorial optimization tasks and solver paradigms. Our goal is to show that \texttt{ACEvo} not only generates challenging instances that expose weaknesses of existing methods, but also evolves solvers that adapt to these harder distributions. We consider three representative routing problems: the Traveling Salesman Problem with Guided Local Search (TSP\_GLS), the Traveling Salesman Problem with Ant Colony Optimization (TSP\_ACO), and the Orienteering Problem with Ant Colony Optimization (OP\_ACO), which together provide a broad testbed for assessing generality. Performance is mainly measured by relative optimality gap, with supplementary analyses including anytime curves, objective values, and instance visualizations. To ensure fairness, we compare against strong baselines such as FunSearch~\cite{RomeraParedes2024MathDiscoveryLLM}, EoH~\cite{LiuICML24}, LLM-LNS~\cite{ye2025large}, KGLS~\cite{arnold2019knowledge} and ReEvo~\cite{YeReEvo24} under both 
standard and \texttt{ACEvo}-generated datasets. Across all tasks, \texttt{ACEvo} consistently induces larger performance gaps for baselines while yielding solvers that close these gaps. Overall, the results confirm that \texttt{ACEvo} produces significantly harder instance distributions while synthesizing solvers with stronger adaptability and generalization. Statistical robustness and sensitivity analysis is provided in Appendix~\ref{app:robustness}.

\subsection{Hardness Amplification and Solver Adaptation under \texttt{ACEvo}}

\paragraph{Baseline vulnerability under \texttt{ACEvo}-generated instances.}
Figure~\ref{fig:tsp_gap_final} shows that \texttt{ACEvo}-generated instances significantly increase the performance gap across all TSP-GLS solvers and problem sizes. While FunSearch~\cite{RomeraParedes2024MathDiscoveryLLM}, EoH~\cite{LiuICML24}, ReEvo~\cite{YeReEvo24}, and KGLS~\cite{arnold2019knowledge} perform well on standard data, their performance drops markedly on \texttt{ACEvo} instances, with the largest gap difference reaching 9\%. These results underscore the limited generalization of existing solvers and validate \texttt{ACEvo}’s ability to expose structural weaknesses through harder instance distributions.

\paragraph{Solver adaptation on hard instance distributions.}
Table~\ref{tab:tsp_gls_gap_table} compares performance gaps across standard and \texttt{ACEvo}-generated hard TSP datasets, evaluated over a wide range of instance sizes (TSP400–TSP1000). All baseline solvers experience a sharp degradation in performance on the hard instances, confirming the increased difficulty induced by \texttt{ACEvo}’s generator. In particular, while the gap values exceed 9\% for the strongest baselines, the \texttt{ACEvo}-generated solver consistently achieves lower gaps, demonstrating its ability to co-adapt to the evolving distribution. These results highlight the dual effectiveness of \texttt{ACEvo} in synthesizing both harder problem instances and stronger, distribution-aware solvers.

\begin{table}[htbp]
\centering
\caption{Performance comparison on standard and proposed hard TSP datasets
(TSP400--TSP1000) in terms of optimality gap (\%). Lower is better.
\textbf{\texttt{ACEvo} (ours)} consistently generalizes to harder instances
while baseline performance degrades.}
\label{tab:tsp_gls_gap_table}

\renewcommand{\arraystretch}{1.25}
\setlength{\tabcolsep}{2.5pt}

\resizebox{0.82\linewidth}{!}{%
\begin{tabular}{lccccccc}
\toprule
\textbf{Method} & \textbf{TSP400} & \textbf{TSP500} &
\textbf{TSP600} & \textbf{TSP700} & \textbf{TSP800} &
\textbf{TSP900} & \textbf{TSP1000} \\
\midrule
\multicolumn{8}{l}{\textit{Standard Dataset}} \\
FunSearch (std) & \textbf{0.738} & \textbf{0.993} & 1.236 & 1.629 & 1.994 & 2.126 & 2.408 \\
EoH (std) & 1.583 & 2.144 & 2.561 & 2.911 & 3.129 & 3.298 & 3.483 \\
LLM-LNS (std) & 0.959 & 1.318 & 1.233 & 1.351 & 1.523 & 1.641 & 1.981 \\
KGLS (std) & 0.854 & 1.105 & 1.248 & 1.349 & 1.500 & 1.599 & \textbf{1.649} \\
ReEvo (std) & 0.806 & 1.034 & 1.249 & 1.377 & 1.518 & 1.623 & 1.724 \\
\textbf{ACEvo (ours)$^\dagger$} &
0.859 & 1.060 & \textbf{1.223} & \textbf{1.319} &
\textbf{1.490} & \textbf{1.565} & 1.656 \\
\midrule
\multicolumn{8}{l}{\textit{\textbf{Proposed Hard Instance Dataset (Ours)}}} \\
\rowcolor{gray!10}
FunSearch (hard) & 9.997 & 13.263 & 8.391 & 9.286 & 10.491 & 9.694 & 9.052 \\
\rowcolor{gray!10}
EoH (hard) & 10.497 & 13.617 & 8.753 & 9.551 & 10.666 & 10.129 & 9.579 \\
\rowcolor{gray!10}
LLM-LNS (hard) & 10.052 & 13.142 & 8.421 & 9.244 & 10.395 & 9.608 & 8.952 \\
\rowcolor{gray!10}
KGLS (hard) & 7.293 & 10.440 & 6.809 & 7.735 & 9.017 & 8.546 & 7.791 \\
\rowcolor{gray!10}
ReEvo (hard) & 7.305 & 10.435 & 6.890 & 7.718 & 9.042 & 8.525 & 7.851 \\
\rowcolor{gray!10}
\textbf{ACEvo (ours)$^\dagger$} &
\textbf{5.484} & \textbf{8.081} & \textbf{5.732} &
\textbf{6.981} & \textbf{8.212} & \textbf{8.185} &
\textbf{7.672} \\
\bottomrule
\end{tabular}%
}

\vspace{2pt}
\footnotesize{$^\dagger$ Our proposed method.}
\end{table}

\FloatBarrier

\definecolor{funblue}{RGB}{52, 152, 219}    
\definecolor{eogreen}{RGB}{26, 188, 156}     
\definecolor{kglsorange}{RGB}{230, 126, 34}  
\definecolor{reevopurple}{RGB}{155, 89, 182} 

\begin{figure*}[htbp]
\centering
\resizebox{\textwidth}{!}{%
\begin{tikzpicture}
  \def\figw{9cm}
  \def\figh{7.2cm}
  \def\fontsz{\Large}
  \def\thick{1.4pt}

  \node[anchor=south] at (0,3.5) {\Large\textbf{FunSearch}};
  \node[anchor=south] at (9,3.5) {\Large\textbf{EoH}};
  \node[anchor=south] at (18,3.5) {\Large\textbf{KGLS}};
  \node[anchor=south] at (27,3.5) {\Large\textbf{ReEvo}};

  \node at (0,0) {
    \begin{tikzpicture}
    \begin{axis}[
        width=\figw, height=\figh,
        xlabel={Problem Size ($n$)},
        ylabel={Gap (\%)},
        xtick={20,100,200,300},
        ymin=0, ymax=10,
        tick label style={font=\fontsz},
        label style={font=\fontsz},
        axis line style={line width=\thick},
        tick style={line width=0.8pt},
        major grid style={line width=0.4pt, draw=gray!30},
        grid=major,
        line width=\thick,
        no markers
    ]
    \addplot[color=funblue, dashed] coordinates {(20,0)(50,0)(100,0.016)(200,0.263)(300,0.576)};
    \addplot[color=funblue, solid] coordinates {(20,0.008)(50,1.477)(100,4.551)(200,7.369)(300,9.142)};
    \end{axis}
    \end{tikzpicture}
  };

  \node at (9,0) {
    \begin{tikzpicture}
    \begin{axis}[
        width=\figw, height=\figh,
        xlabel={Problem Size ($n$)},
        xtick={20,100,200,300},
        ymin=0, ymax=10,
        tick label style={font=\fontsz},
        label style={font=\fontsz},
        axis line style={line width=\thick},
        tick style={line width=0.8pt},
        major grid style={line width=0.4pt, draw=gray!30},
        grid=major,
        line width=\thick,
        no markers
    ]
    \addplot[color=eogreen, dashed] coordinates {(20,0)(50,0)(100,0.037)(200,0.344)(300,0.881)};
    \addplot[color=eogreen, solid] coordinates {(20,0)(50,1.777)(100,5.039)(200,8.259)(300,9.910)};
    \end{axis}
    \end{tikzpicture}
  };

  \node at (18,0) {
    \begin{tikzpicture}
    \begin{axis}[
        width=\figw, height=\figh,
        xlabel={Problem Size ($n$)},
        xtick={20,100,200,300},
        ymin=0, ymax=10,
        tick label style={font=\fontsz},
        label style={font=\fontsz},
        axis line style={line width=\thick},
        tick style={line width=0.8pt},
        major grid style={line width=0.4pt, draw=gray!30},
        grid=major,
        line width=\thick,
        no markers
    ]
    \addplot[color=kglsorange, dashed] coordinates {(20,0.004)(50,0.017)(100,0.010)(200,0.284)(300,0.584)};
    \addplot[color=kglsorange, solid] coordinates {(20,0.027)(50,0.422)(100,1.083)(200,3.642)(300,5.465)};
    \end{axis}
    \end{tikzpicture}
  };

  \node at (27,0) {
    \begin{tikzpicture}
    \begin{axis}[
        width=\figw, height=\figh,
        xlabel={Problem Size ($n$)},
        xtick={20,100,200,300},
        ymin=0, ymax=10,
        tick label style={font=\fontsz},
        label style={font=\fontsz},
        axis line style={line width=\thick},
        tick style={line width=0.8pt},
        major grid style={line width=0.4pt, draw=gray!30},
        grid=major,
        line width=\thick,
        no markers
    ]
    \addplot[color=reevopurple, dashed] coordinates {(20,0)(50,0)(100,0.011)(200,0.216)(300,0.548)};
    \addplot[color=reevopurple, solid] coordinates {(20,0.010)(50,0.500)(100,1.114)(200,3.562)(300,5.596)};
    \end{axis}
    \end{tikzpicture}
  };

\end{tikzpicture}
}
\vspace{-0.6em}
\caption{\small Gap (\%) comparison across four methods on \textbf{standard (dashed)} and \textbf{hard (solid)} datasets. Lower values indicate better performance, while higher values reflect greater degradation under hard instance distributions.}
\label{fig:tsp_gap_final}
\end{figure*}
\FloatBarrier  
\vspace{-1em}

\paragraph{Solver--instance synergy on the Orienteering Problem.}
Table~\ref{tab:op_aco_obj} reports the performance of different heuristic--instance combinations for the Orienteering Problem solved via Ant Colony Optimization (OP\_ACO). While heuristics generated by ReEvo~\cite{YeReEvo24} remain reasonably effective on the hard instances constructed by \texttt{ACEvo}, the heuristics co-evolved within the \texttt{ACEvo} framework consistently outperform them across all evaluated problem sizes, indicating that jointly evolving heuristics and instances is more effective than relying on pre-trained or manually designed solvers. The heuristics synthesized by \texttt{ACEvo} require neither hand-crafted rule modifications nor hyperparameter tuning; instead, they are generated through symbolic interactions with large language models and refined via evolutionary reflection that integrates behavioral feedback into subsequent prompts. In contrast, static heuristics often generalize poorly under distribution shift, whereas the co-evolutionary dynamics of \texttt{ACEvo} enable solvers and instances to adapt to each other, allowing the automatic construction of challenging instance distributions and the discovery of robust optimization strategies for dynamic environments.
Consistent solver--instance synergy is also observed for the Capacitated Vehicle Routing Problem solved via Ant Colony Optimization (CVRP\_ACO); detailed results are reported in Appendix~\ref{appendix:cvrp}.

\paragraph{Instance-induced degradation under fixed solvers.}
To evaluate the intrinsic hardness of \texttt{ACEvo}-generated instances independently of solver adaptation, 
we assess the performance of ReEvo’s~\cite{YeReEvo24} heuristics on three representative combinatorial optimization tasks:
the Traveling Salesman Problem solved via Ant Colony Optimization (TSP\_ACO), 
the Orienteering Problem solved via Ant Colony Optimization (OP\_ACO), 
and the Capacitated Vehicle Routing Problem solved via Ant Colony Optimization (CVRP\_ACO).
Figure~\ref{fig:tsp_op_cvrp_aco_objectives} reports the objective values obtained on both standard and 
\texttt{ACEvo}-generated instance distributions across a range of problem sizes.
In the TSP and CVRP settings, higher objective values correspond to worse solutions, whereas in OP, lower objective values indicate reduced reward collection.
Across all three tasks, \texttt{ACEvo}-generated instances consistently lead to more unfavorable objective outcomes,
demonstrating that the induced instance distributions effectively expose solver weaknesses
under static, non-adaptive evaluation.

\definecolor{reevoSoft}{RGB}{120,170,220}  
\definecolor{ealgSoft}{RGB}{210,120,120}   

\pgfplotsset{
  compat=1.18,
  every axis/.append style={
    legend style={font=\small},
    legend to name=sharedlegend
  }
}

\begin{figure*}[t]
\centering

\begin{minipage}[t]{0.32\textwidth}
\centering
\begin{tikzpicture}
\begin{axis}[
    title={TSP\_ACO},
    width=0.95\textwidth,
    height=4.0cm,
    xlabel={Problem Size ($n$)},
    ylabel={Objective Value},
    grid=major,
    xtick={400,600,800,1000},
    tick label style={font=\footnotesize},
    yticklabel style={font=\footnotesize,text width=2.2em,align=right},
    label style={font=\footnotesize},
    title style={font=\footnotesize},
    every axis plot/.append style={thick},
    ymin=17
]
\addplot+[
    mark=*,
    smooth,
    color=reevoSoft
] coordinates {
    (400, 17.06) (500, 19.58) (600, 21.89) (700, 24.05)
    (800, 26.07) (900, 28.09) (1000, 29.95)
};
\addlegendentry{ReEvo $\oplus$ Std Obj}

\addplot+[
    mark=square*,
    color=ealgSoft
] coordinates {
    (400, 19.19) (500, 22.80) (600, 30.79) (700, 29.32)
    (800, 39.17) (900, 39.69) (1000, 45.65)
};
\addlegendentry{ReEvo $\oplus$ Hard(ACEvo) Obj}
\end{axis}
\end{tikzpicture}
\end{minipage}%
\hfill
\begin{minipage}[t]{0.32\textwidth}
\centering
\begin{tikzpicture}
\begin{axis}[
    title={OP\_ACO},
    width=0.95\textwidth,
    height=4.0cm,
    xlabel={Problem Size ($n$)},
    ylabel={Objective Value},
    grid=major,
    xtick={400,600,800,1000},
    tick label style={font=\footnotesize},
    yticklabel style={font=\footnotesize,text width=2.2em,align=right},
    label style={font=\footnotesize},
    title style={font=\footnotesize},
    every axis plot/.append style={thick},
    ymin=0
]
\addplot+[
    mark=*,
    smooth,
    color=reevoSoft
] coordinates {
    (400, 112.25) (500, 128.31) (600, 139.87) (700, 151.19)
    (800, 162.04) (900, 171.26) (1000, 182.60)
};
\addlegendentry{ReEvo $\oplus$ Std Obj}

\addplot+[
    mark=square*,
    color=ealgSoft
] coordinates {
    (400, 17.77) (500, 24.94) (600, 22.65) (700, 16.46)
    (800, 23.21) (900, 20.22) (1000, 20.06)
};
\addlegendentry{ReEvo $\oplus$ Hard(ACEvo) Obj}
\end{axis}
\end{tikzpicture}
\end{minipage}%
\hfill
\begin{minipage}[t]{0.32\textwidth}
\centering
\begin{tikzpicture}
\begin{axis}[
    title={CVRP\_ACO},
    width=0.95\textwidth,
    height=4.0cm,
    xlabel={Problem Size ($n$)},
    ylabel={Objective Value},
    grid=major,
    xtick={400,600,800,1000},
    tick label style={font=\footnotesize},
    yticklabel style={font=\footnotesize,text width=2.2em,align=right},
    label style={font=\footnotesize},
    title style={font=\footnotesize},
    every axis plot/.append style={thick},
    ymin=50
]
\addplot+[
    mark=*,
    smooth,
    color=reevoSoft
] coordinates {
    (400, 54.3703945743559)
    (500, 65.82010872319373)
    (600, 77.19005913243943)
    (700, 87.54118026840491)
    (800, 98.15107475268942)
    (900, 108.91908490447096)
    (1000, 118.69600549581716)
};
\addlegendentry{ReEvo $\oplus$ Std Obj}

\addplot+[
    mark=square*,
    color=ealgSoft
] coordinates {
    (400, 59.35069059464591)
    (500, 73.44901762774533)
    (600, 87.29248794794175)
    (700, 100.85927531435544)
    (800, 114.31035825447456)
    (900, 128.94379077695496)
    (1000, 142.15134067790117)
};
\addlegendentry{ReEvo $\oplus$ Hard(ACEvo) Obj}
\end{axis}
\end{tikzpicture}
\end{minipage}

\vspace{0.3em}

\vspace{-0.5em}
\caption{
\textbf{Blue line:} performance on standard instances;
\textbf{Red line:} performance on \texttt{ACEvo}-generated hard instances.
Objective values using ReEvo heuristics on \textbf{TSP\_ACO (left)},
\textbf{OP\_ACO (middle)}, and \textbf{CVRP\_ACO (right)}.
In TSP and CVRP, higher objective values indicate worse solutions, whereas in OP,
lower objective values indicate worse solutions.
Across tasks, \texttt{ACEvo}-generated instances consistently induce more adverse outcomes,
indicating increased optimization difficulty.
}
\label{fig:tsp_op_cvrp_aco_objectives}
\vspace{-1.5em}
\end{figure*}
\FloatBarrier

\subsection{Generalization for the Traveling Salesman Problem}
To further assess the generalization capability of \texttt{ACEvo}-generated solvers, we conduct evaluations on classical reference instances from the TSPLIB~\cite{Reinelt91} dataset, a widely recognized benchmark for the Traveling Salesman Problem (TSP). These instances span diverse structural characteristics and problem scales, reflecting the complexity of real-world routing scenarios. By incorporating TSPLIB~\cite{Reinelt91}, we examine whether solvers evolved under \texttt{ACEvo} transfer effectively beyond the synthetic distributions used during training.

We compare \texttt{ACEvo} against three competitive LLM-based frameworks—FunSearch~\cite{RomeraParedes2024MathDiscoveryLLM}, ReEvo~\cite{YeReEvo24}, and LLM-LNS~\cite{ye2025large}. Across all evaluated TSPLIB instances with up to 2000 nodes, using Guided Local Search (GLS) as the underlying solver, \texttt{ACEvo} consistently achieves the lowest optimality gap, indicating superior adaptability and robustness.

These results suggest that the co-evolutionary process in \texttt{ACEvo} yields solvers that generalize beyond synthetic hard instances and remain effective on human-designed benchmarks. The consistent improvements over strong baselines further highlight the role of adversarial feedback in achieving robust generalization. The complete TSPLIB~\cite{Reinelt91} results are reported in Table~\ref{tab:tsplib_full_single}.

\subsection{Ablation Study on Evolutionary Components}

As detailed in Appendix~\ref{app:ablation_components}, we analyze the contribution of individual evolutionary operators in \texttt{ACEvo} by disabling adversarial feedback, reflection, or mutation, and evaluating the resulting solver gaps on hard instances.
Removing \emph{adversarial feedback} leads to the largest degradation in solver performance, with the solver gap increasing to 11.43\% despite the generated instances being less challenging (7.76\%), indicating that reduced instance difficulty alone does not yield stronger solvers and that adaptive pressure from adversarial interaction is critical for effective solver evolution.
Disabling \emph{reflection} increases the solver gap to 10.21\%, as the absence of structured diagnostic signals results in less targeted solver adaptation and weaker performance gains.
When \emph{mutation} is removed, the solver gap rises to 9.33\%; although its effect is smaller than that of adversarial feedback or reflection, mutation remains important for maintaining population diversity and preventing evolutionary stagnation.
With all components enabled, the full \texttt{ACEvo} achieves the lowest solver gap (8.10\%) while generating the hardest instances (10.43\%), showing that adversarial feedback, reflection, and mutation play complementary roles and must be jointly applied to support stable co-evolution and robust solver generalization.

\begin{table*}[t]

\captionsetup{justification=raggedright,singlelinecheck=false}
\caption{Comprehensive evaluation of \texttt{ACEvo}$^{\dagger}$ and baseline frameworks on TSPLIB benchmark instances.
Reported values indicate the optimality gap (\%) relative to known optima.
Bold numbers highlight results achieved by \texttt{ACEvo}, indicating cases where it attains or matches the best performance.
$^{\dagger}$\,denotes our proposed method.}

\label{tab:tsplib_full_single}
\centering

\small
\setlength{\tabcolsep}{3pt}
\renewcommand{\arraystretch}{1.04}
\setlength{\arrayrulewidth}{0.5pt}

\resizebox{\textwidth}{!}{%
\begin{tabular}{
l c c c c c
!{\vrule width 0.5pt}
l c c c c c
!{\vrule width 0.5pt}
l c c c c c
}
\hline

\textbf{Instance} & {FunSearch} & {EoH} & {LLM--LNS} & {ReEvo} &
{\textbf{\texttt{ACEvo}}$^{\dagger}$} &
\textbf{Instance} & {FunSearch} & {EoH} & {LLM--LNS} & {ReEvo} &
{\textbf{\texttt{ACEvo}}$^{\dagger}$} &
\textbf{Instance} & {FunSearch} & {EoH} & {LLM--LNS} & {ReEvo} &
{\textbf{\texttt{ACEvo}}$^{\dagger}$} \\

\hline

kroE100   & 0.07 & 0.00 & 0.00 & 0.00 & \textbf{0.00} &
pr124     & 0.00 & 0.00 & 0.00 & 0.00 & \textbf{0.00} &
rd400     & 0.76 & 2.23 & 0.82 & 1.11 & 0.82 \\

pr226     & 0.12 & 0.10 & 0.06 & 0.00 & \textbf{0.00} &
kroB150   & 0.03 & 0.08 & 0.01 & 0.00 & \textbf{0.00} &
u724      & 1.81 & 2.85 & 1.13 & 1.17 & \textbf{1.02} \\

vm1748    & 3.88 & 4.33 & 4.33 & 1.57 & \textbf{1.14} &
ulysses22 & 0.12 & 0.00 & 0.00 & 0.00 & \textbf{0.00} &
kroB100   & 0.00 & 0.00 & 0.00 & 0.00 & \textbf{0.00} \\

pr1002    & 1.89 & 3.27 & 1.16 & 0.75 & \textbf{0.57} &
bier127   & 0.12 & 0.26 & 0.01 & 0.16 & \textbf{0.01} &
kroD100   & 0.00 & 0.00 & 0.00 & 0.00 & \textbf{0.00} \\

pr152     & 0.00 & 0.00 & 0.19 & 0.19 & \textbf{0.00} &
kroA100   & 0.02 & 0.02 & 0.02 & 0.02 & \textbf{0.02} &
u159      & 0.00 & 0.00 & 0.00 & 0.00 & \textbf{0.00} \\

dsj1000   & 3.43 & 4.28 & 1.06 & 0.74 & \textbf{0.73} &
fl1577    & 4.79 & 5.03 & 5.03 & 2.34 & \textbf{1.61} &
st70      & 0.31 & 0.31 & 0.31 & 0.31 & \textbf{0.31} \\

rat195    & 0.96 & 0.99 & 1.37 & 1.38 & \textbf{0.96} &
berlin52  & 0.03 & 0.03 & 0.03 & 0.03 & \textbf{0.03} &
kroA200   & 0.13 & 0.25 & 0.62 & 0.05 & \textbf{0.05} \\

pr136     & 0.04 & 0.09 & 0.00 & 0.12 & \textbf{0.00} &
u1060     & 2.66 & 4.04 & 1.54 & 1.33 & \textbf{1.14} &
rat99     & 0.68 & 0.68 & 0.68 & 0.68 & \textbf{0.68} \\

pr264     & 1.01 & 0.00 & 0.00 & 1.01 & 1.01 &
kroA150   & 0.01 & 0.00 & 0.00 & 0.00 & \textbf{0.00} &
lin318    & 0.98 & 1.46 & 1.09 & 0.70 & \textbf{0.70} \\

d493      & 1.51 & 2.82 & 1.27 & 0.83 & \textbf{0.72} &
ch130     & 0.02 & 0.01 & 0.70 & 0.01 & \textbf{0.01} &
fl3795    & 6.55 & 4.38 & 4.38 & 2.69 & \textbf{2.40} \\

pr76      & 0.00 & 0.00 & 0.00 & 0.00 & \textbf{0.00} &
pcb1173   & 3.91 & 5.07 & 2.91 & 1.13 & \textbf{0.92} &
u1432     & 4.37 & 4.84 & 3.02 & 1.18 & \textbf{1.07} \\

a280      & 0.92 & 2.06 & 0.34 & 0.30 & 0.66 &
fl1400    & 4.88 & 7.66 & 2.28 & 3.07 & \textbf{2.28} &
ts225     & 0.00 & 0.00 & 0.00 & 0.00 & \textbf{0.00} \\

kroB200   & 0.15 & 0.23 & 0.44 & 0.04 & \textbf{0.04} &
d198      & 0.46 & 0.40 & 0.29 & 0.31 & 0.40 &
u1817     & 4.78 & 4.62 & 4.62 & 2.34 & \textbf{2.00} \\

nrw1379   & 3.48 & 3.82 & 2.99 & 1.79 & \textbf{1.12} &
d657      & 1.66 & 2.85 & 1.02 & 0.82 & \textbf{0.72} &
ch150     & 0.12 & 0.37 & 0.04 & 0.04 & \textbf{0.04} \\

lin105    & 2.77 & 0.03 & 0.03 & 0.03 & \textbf{0.03} &
rl1889    & 4.86 & 4.08 & 4.08 & 1.50 & \textbf{1.20} &
d1655     & 5.04 & 5.79 & 5.79 & 2.42 & \textbf{2.31} \\

tsp225    & 0.63 & 1.39 & 0.00 & 0.00 & \textbf{0.00} &
rat783    & 2.86 & 4.48 & 2.18 & 1.66 & \textbf{1.22} &
pr107     & 0.00 & 0.00 & 0.00 & 0.00 & \textbf{0.00} \\

fl417     & 0.63 & 0.80 & 0.77 & 0.60 & \textbf{0.58} &
p654      & 0.22 & 0.75 & 0.05 & 0.16 & \textbf{0.03} &
vm1084    & 1.68 & 3.64 & 1.74 & 0.45 & 0.61 \\

rd100     & 0.01 & 0.01 & 0.01 & 0.01 & \textbf{0.01} &
rl1323    & 3.29 & 4.35 & 1.93 & 2.00 & \textbf{1.71} &
linhp318  & 2.59 & 3.22 & 2.77 & 2.28 & \textbf{1.99} \\

pcb442    & 0.84 & 1.15 & 0.96 & 1.06 & \textbf{0.53} &
eil101    & 1.94 & 2.59 & 2.08 & 1.78 & \textbf{1.78} &
rat575    & 2.33 & 3.11 & 1.88 & 1.28 & \textbf{1.21} \\

gil262    & 0.53 & 0.59 & 0.48 & 0.40 & \textbf{0.37} &
eil76     & 1.23 & 1.53 & 1.18 & 1.18 & \textbf{1.18} &
pr299     & 0.25 & 0.61 & 0.11 & 0.21 & 0.16 \\

pr439     & 1.56 & 2.80 & 1.97 & 0.13 & 0.60 &
kroC100   & 0.01 & 0.01 & 0.01 & 0.01 & \textbf{0.01} &
eil51     & 0.67 & 0.67 & 0.67 & 0.67 & \textbf{0.67} \\

\hline
\end{tabular}%
}

\end{table*}

\FloatBarrier






\subsection{Structural Visualization of Optimization Instances}

To gain qualitative insights into the structural properties of \texttt{ACEvo}-generated instances, we visualize the spatial distributions of four representative problem types—TSP\_GLS, TSP\_ACO, OP\_ACO ,and CVRP\_ACO—using two nonlinear manifold learning techniques, t-SNE~\cite{van2008visualizing} and UMAP~\cite{2018arXivUMAP}, as well as standard coordinate plots (Appendix~\ref{appendix:vis_protocol}). Each visualization is performed over a batch of 500 instances. Across all tasks, the standard (bottom row) instance distributions exhibit uniform dispersion and low geometric complexity, consistent with conventional Euclidean instance distributions. In contrast, \texttt{ACEvo}-generated hard instances (top and middle rows) reveal pronounced structural patterns: t-SNE projections expose fragmented clusters and boundary concentrations, while UMAP highlights manifold separation and high-density cores. These effects are particularly evident in OP\_ACO, where diverse structural motifs emerge, and in CVRP\_ACO, which exhibits anisotropic and curved manifolds reflecting capacity-induced coupling and depot-centered routing constraints. Taken together, the above observations indicate that \texttt{ACEvo} not only increases algorithmic difficulty but also fundamentally reshapes the instance distribution at the distributional level; these changes are clearly discernible in both visual and structural characteristics and further amplify the combinatorial complexity of the resulting optimization problems.

\section{Conclusion}

We introduce \texttt{ACEvo}, a combinatorial optimization framework that enables adversarial co-evolution of problem instances and solvers mediated by large language models. By replacing static instance distributions and manually designed heuristics with a feedback-driven evolutionary loop, \texttt{ACEvo} jointly amplifies instance hardness and solver capability, enabling the autonomous emergence of adaptive algorithms. Experiments on classical combinatorial optimization tasks, including the Traveling Salesman Problem (TSP), Orienteering Problem (OP), and Capacitated Vehicle Routing Problem (CVRP), across widely used public datasets and TSPLIB instances demonstrate that \texttt{ACEvo} generates substantially harder instances while achieving state-of-the-art performance, with synthesized solvers outperforming leading LLM-based baselines. \texttt{ACEvo} opens a new frontier for research at the intersection of adversarial instance generation, solver synthesis, and combinatorial optimization. By coupling evolutionary co-evolution with language-model reasoning, it establishes a pathway toward sustained, self-driven algorithmic innovation.

\section*{Acknowledgments}
This work was supported by the Institute of Information and Communications Technology Planning and Evaluation (IITP) grant funded by the Korean government (MSIT) (No. RS-2025-02214654), and by the New Faculty Startup Fund from Seoul National University.


\bibliographystyle{colm2026_conference}
\bibliography{colm2026_conference}

\newpage
\appendix
\onecolumn

\clearpage
\appendix

\section*{\centering\large Appendix Overview}
\addcontentsline{toc}{section}{Appendix Overview}

\begingroup
\setlength{\parindent}{0pt}
\normalsize
\centering

\vspace{0.4em}
\hrule
\vspace{0.7em}

{\itshape
This appendix provides extended theoretical formulations, prompt specifications,
experimental configurations, and supplementary empirical analyses supporting the main paper.
}

\vspace{1.0em}

\newcommand{\appmain}[2]{%
  \vspace{0.6em}
  \noindent\textbf{#1}%
  \leaders\hbox to 0.6em{\hss.\hss}\hfill \pageref{#2}\par
}

\newcommand{\appsub}[2]{%
  \noindent\hspace*{1.5em}{\footnotesize #1}%
  \leaders\hbox to 0.6em{\hss.\hss}\hfill \pageref{#2}\par
}

\appmain{A. Extended Preliminaries}{app:ext-prelim}
\appsub{A.1 Formal Problem Setting}{app:formal_problem}
\appsub{A.2 Solver Evaluation over Generated Instance Distributions}{app:solver_eval}
\appsub{A.3 Background on the Traveling Salesman Problem}{app:tsp_background}
\appsub{A.4 Background on the Orienteering Problem}{app:op_background}
\appsub{A.5 Background on the Capacitated Vehicle Routing Problem}{app:cvrp_background}
\appsub{A.6 Heuristic Solvers for Routing Problems}{app:routing_heuristics}
\appsub{A.7 Instance Distribution Shift in Combinatorial Optimization}{app:distribution_shift}
\appsub{A.8 Objective Direction and Hardness Signals}{app:objective_hardness}
\appsub{A.9 Primary Prompts}{app:primary_prompts}
\appsub{A.10 Task-Tailored Prompt Components}{app:task_tailored_prompts}

\appmain{B. Extended Applications and Generalization Scope}{appendix:extended_applications}

\appmain{C. Experimental Details}{app:exp_details}
\appsub{C.1 Guided Local Search for the Traveling Salesman Problem}{app:gls_tsp}
\appsub{C.2 Experimental Configuration}{app:exp_config}
\appsub{C.3 Objective Value Definition in TSP ACO, OP ACO, and CVRP ACO}{app:objective_definition}

\appmain{D. Detailed Analysis}{app:detailed_analysis}
\appsub{D.1 Detailed Description of the Visualization Protocol}{appendix:vis_protocol}
\appsub{D.2 Evolution Process}{subsec:evolution-process}
\appsub{D.3 Statistical Robustness and Sensitivity Analysis}{app:robustness}
\appsub{D.4 Solver--instance Synergy on the Orienteering Problem}{appendix:op}
\appsub{D.5 Solver--instance Synergy on the Capacitated Vehicle Routing Problem}{appendix:cvrp}
\appsub{D.6 Ablation Study on Evolutionary Components}{app:ablation_components}
\appsub{D.7 Adversarial Co-Evolution Procedure of \texttt{ACEvo}}{app:acevo-pseudocode}

\vspace{0.8em}
\hrule

\endgroup

\newpage
\section{Extended Preliminaries}
\label{app:ext-prelim}

\subsection{Formal Problem Setting}
\label{app:formal_problem}
Let $\mathcal{I}$ denote the space of problem instances and $\mathcal{S}$ the space of feasible solutions. 
A combinatorial optimization problem is specified by an objective function
\begin{equation}
f : \mathcal{S} \rightarrow \mathbb{R},
\end{equation}
where lower values correspond to better solutions.

A heuristic solver is modeled as a mapping
\begin{equation}
h : \mathcal{I} \rightarrow \mathcal{S}, \quad h \in \mathcal{H},
\end{equation}
where $\mathcal{H}$ denotes the hypothesis space of admissible solvers.

An instance generator is defined as a mapping
\begin{equation}
g : \mathcal{Z} \rightarrow \mathcal{I}, \quad g \in \mathcal{G},
\end{equation}
where $\mathcal{Z}$ is a latent seed space equipped with a base distribution $z \sim \mathcal{Z}$.

The generator $g$ induces a distribution over problem instances:
\begin{equation}
I \sim \mathcal{D}_g, 
\qquad 
\mathcal{D}_g := g_{\#}(\mathcal{Z}),
\end{equation}
where $g_{\#}$ denotes the pushforward measure of $g$.

\subsection{Solver Evaluation over Generated Instance Distributions}
\label{app:solver_eval}

Given a solver $h \in \mathcal{H}$ and a generator $g \in \mathcal{G}$, solver performance is evaluated in expectation over the induced instance distribution:
\begin{equation}
\mathcal{L}(h; g)
\;=\;
\mathbb{E}_{I \sim \mathcal{D}_g}
\left[
f\big(h(I)\big)
\right].
\end{equation}

In practice, this expectation is approximated via Monte Carlo sampling:
\begin{equation}
\widehat{\mathcal{L}}(h; g)
\;=\;
\frac{1}{n}
\sum_{i=1}^{n}
f\!\left(
h\!\left(g(z_i)\right)
\right),
\qquad
z_i \sim \mathcal{Z}.
\end{equation}

This formulation treats the solver as a black-box mapping from instances to solutions and does not assume differentiability or access to internal solver states.

\subsection{Background on the Traveling Salesman Problem}
\label{app:tsp_background}
The Traveling Salesman Problem (TSP) is a classical routing problem defined on a complete weighted graph. Given a graph
$G = (V, E, w)$, where $|V| = n$, the weight function $w : E \rightarrow \mathbb{R}_{+}$ assigns a non-negative travel cost to each edge.

A feasible solution corresponds to a tour $\pi$ over $V$. The objective is to minimize the total travel cost:
\begin{equation}
\min_{\pi} \sum_{(i,j)\in \pi} w(i,j)
\quad \text{s.t.} \quad
\pi = (v_{\pi_1}, \dots, v_{\pi_n}), \;
\{\pi_1, \dots, \pi_n\} = V.
\end{equation}

\subsection{Background on the Orienteering Problem}
\label{app:op_background}
The Orienteering Problem (OP) is a routing problem that generalizes the Traveling Salesman Problem by associating each node with a non-negative reward and introducing a global budget constraint. Given a complete weighted graph
$G = (V, E, w)$, each node $v \in V$ is assigned a reward $r(v) \ge 0$, and each edge $(i,j) \in E$ has an associated travel cost $w(i,j)$.

A feasible solution corresponds to a tour $\pi \subseteq V$ that starts and ends at a designated depot. The objective is to maximize the total collected reward:
\begin{equation}
\max_{\pi \subseteq V} \sum_{v \in \pi} r(v)
\quad \text{s.t.} \quad
\sum_{(i,j)\in \pi} w(i,j) \le B.
\end{equation}

\subsection{Background on the Capacitated Vehicle Routing Problem}
\label{app:cvrp_background}
The Capacitated Vehicle Routing Problem (CVRP) extends classical routing problems by introducing multiple routes and explicit capacity constraints. Given a set of customer nodes $V = \{1,\dots,n\}$ and a depot node $0$, each customer $i \in V$ is associated with a demand $d_i > 0$. A fleet of identical vehicles, each with capacity $Q$, departs from and returns to the depot to serve all customers.

A feasible solution consists of a set of routes $\{\pi_1, \dots, \pi_K\}$. The objective is to minimize the total routing cost:
\begin{equation}
\min \sum_{k=1}^{K} \sum_{(i,j)\in \pi_k} w(i,j)
\quad \text{s.t.} \quad
\sum_{i \in \pi_k} d_i \le Q, \quad \forall k = 1,\dots,K.
\end{equation}

\subsection{Heuristic Solvers for Routing Problems}
\label{app:routing_heuristics}
Heuristic solvers for routing problems can be abstracted as iterative procedures that refine an initial feasible solution
\begin{equation}
\pi^{(0)} \in \mathcal{S}(I)
\end{equation}
through a sequence of solver-dependent transformations:
\begin{equation}
\pi^{(t+1)} = \mathcal{T}_{\theta}\!\left(\pi^{(t)}, I\right),
\end{equation}
where $\mathcal{T}_{\theta}$ denotes a heuristic update operator parameterized by solver-specific logic $\theta$, and $I$ represents the problem instance.

This abstraction encompasses a wide range of classical heuristics, including local-search-based methods and population-based approaches. Different solvers instantiate $\mathcal{T}_{\theta}$ using distinct mechanisms, while sharing the common objective of iteratively improving solution quality under the constraints imposed by the instance structure.

\subsection{Instance Distribution Shift in Combinatorial Optimization}
\label{app:distribution_shift}
Let $\mathcal{D}_{\mathrm{std}}$ denote a reference instance distribution and $\mathcal{D}_g$ an instance distribution induced by a generator $g \in \mathcal{G}$. A distribution shift occurs when
\begin{equation}
\mathcal{D}_g \neq \mathcal{D}_{\mathrm{std}}.
\end{equation}

Solver behavior under distributional shift can be characterized by comparing expected performance across distributions:
\begin{equation}
\mathbb{E}_{I \sim \mathcal{D}_{\mathrm{std}}} \!\left[ f\!\left(h(I)\right) \right]
\quad \text{and} \quad
\mathbb{E}_{I \sim \mathcal{D}_g} \!\left[ f\!\left(h(I)\right) \right],
\end{equation}
where $h$ denotes a solver and $f(\cdot)$ the associated objective function.

By iteratively mutating generator programs in $\mathcal{G}$, \texttt{ACEvo} induces a sequence of instance distributions
\begin{equation}
\mathcal{D}_{g_1}, \mathcal{D}_{g_2}, \dots, \mathcal{D}_{g_T},
\end{equation}
which enables the systematic study of solver behavior under progressively shifting instance distributions.

\subsection{Objective Direction and Hardness Signals}
\label{app:objective_hardness}
The routing problems considered in this work include both minimization and maximization objectives. Specifically, TSP and CVRP are formulated as minimization problems, whereas OP is formulated as a maximization problem. For notational consistency, solver performance is evaluated directly using the task-specific objective function $f(\cdot)$.

When assessing instance difficulty under a fixed solver, degradation in solver performance manifests differently depending on the objective direction. For minimization problems, larger objective values indicate worse solutions, whereas for maximization problems, smaller objective values indicate worse solutions. Accordingly, instance hardness can be characterized by relative deterioration in solver performance with respect to the reference instance distribution.

\subsection{Primary Prompts}
\label{app:primary_prompts}

\definecolor{lavendermist}{RGB}{243, 238, 252}
\begin{tcolorbox}[colframe=black, colback=lavendermist,  boxrule=0.5pt, arc=5pt]

You are a specialist in algorithmic optimization, with a focus on heuristic strategies. Your task is to construct effective heuristics that leverage underlying structural patterns within the problem to solve optimization tasks. The approach should demonstrate clear improvements in efficiency and solution quality compared to prior methods. 

\end{tcolorbox}

\begin{center}
Prompt 1: System prompt for heuristics generator LLM.
\end{center}

\definecolor{mistblue}{RGB}{225, 235, 245}
\begin{tcolorbox}[colframe=black, colback=lavendermist,  boxrule=0.5pt, arc=5pt]

You are a specialist in generating challenging input distributions for combinatorial optimization problems. Your task is to construct complex problem instances characterized by rich structural diversity. The resulting instances should surpass prior designs in both difficulty and diversity.

\end{tcolorbox}

\begin{center}
Prompt 2: System prompt for hard instance generator LLM.
\end{center}

\definecolor{mistblue}{RGB}{225, 235, 245}
\begin{tcolorbox}[colframe=black, colback=lavendermist,  boxrule=0.5pt, arc=5pt]

Implement the \textbraceleft function\_name\textbraceright\ function to solve \textbraceleft problem\_description\textbraceright.

\end{tcolorbox}

\begin{center}
Prompt 3: Problem Statement.
\end{center}

\definecolor{mistblue}{RGB}{225, 235, 245}
\begin{tcolorbox}[colframe=black, colback=lavendermist,  boxrule=0.5pt, arc=5pt]
{\{problem\_statement\}} \\
\medskip

{\{base\_function\}} \\
\medskip

Invent a creative variation named {\{func\_name\_v2\}}. Return only the code, enclosed in a properly formatted Python code string block like this:
\begin{verbatim}
```python ...```
\end{verbatim}
\end{tcolorbox}

\begin{center}
Prompt 4: Initialization.
\end{center}
\vspace{-0.5em}
\definecolor{mistblue}{RGB}{225, 235, 245}
\begin{tcolorbox}[colframe=black, colback=lavendermist, boxrule=0.5pt, arc=5pt]

{\{problem\_statement\}} \\
\medskip

{\{base\_function\}} \\
\medskip

{[Prior reflection]} \\
\{reflection\} \\
\medskip

{[Previous result]} \\
\{result\} \\
\medskip

{[Code]} \\
\{func\_signature\} \\
\medskip

{[Improved code]} \\
Construct a mutated version of the original function titled \texttt{\{func\_name\}\_v2}, guided by the given feedback. Return only the function’s implementation as a Python string enclosed within a properly formatted code block:\begin{verbatim}
```python ...```
\end{verbatim}

\end{tcolorbox}

{\centering Prompt 5: Mutation (heuristics) or mutation (hard instance)\par}

\definecolor{mistblue}{RGB}{225, 235, 245}
\begin{tcolorbox}[colframe=black, colback=lavendermist, boxrule=0.5pt, arc=5pt,]

[Information]\\
\{mutation\} \\
\medskip

[Adaptivity node] \\
\{diagnostic anchor\}
\end{tcolorbox}

{\centering Prompt 6:  User prompt for LLM.\par}

\subsection{Task-Tailored Prompt Components}
\label{app:task_tailored_prompts}
\definecolor{mistblue}{RGB}{225, 235, 245}
\begin{tcolorbox}[colframe=black, colback=lavendermist, boxrule=0.5pt, arc=5pt]

\#heuristics \\
The "heuristics" function receives a distance matrix as input and outputs a matrix of the same dimensions, where each entry reflects a preliminary estimate of the cost or undesirability of selecting the corresponding edge in a solution.\\
\medskip

\#hard\_instance\_generator\\
The "hard\_instance\_generator" function takes as input a solver, and returns a distance matrix that is adversarially constructed to maximize the solver’s failure or suboptimality. The return is a valid TSP instance in the standard distance matrix format.\\

\end{tcolorbox}

{\centering Prompt 7:  Problem description (TSP\_GLS).\par}

\begin{tcolorbox}[
    colframe=black,
    colback=lavendermist,
    boxrule=0.5pt,
    arc=5pt,
    before skip=6pt,
    after skip=6pt
]

\#heuristics \\
The "heuristics" function processes a given distance matrix and outputs a matrix of the same shape, where each entry reflects a prior estimate of how beneficial it may be to include the corresponding edge in the final solution.\\
\medskip

\#hard\_instance\_generator\\
The "hard\_instance\_generator" function returns a distance matrix of fixed shape, intentionally constructed to be difficult for optimization algorithms to solve, often by embedding deceptive or irregular structural patterns.\\

\end{tcolorbox}
\vspace{-0.5em}
\begin{center}
Prompt 8:  Problem description (TSP\_ACO).
\end{center}
\vspace{-0.5em}
\definecolor{pygKeyword}{RGB}{0,128,0}
\definecolor{pygFunction}{RGB}{0,0,255}
\definecolor{pygComment}{RGB}{61,123,123}
\definecolor{pygString}{RGB}{186,33,33}
\lstdefinestyle{mypython}{
    language=Python,
    basicstyle=\ttfamily\small,
    keywordstyle=\color{pygKeyword}\bfseries,
    commentstyle=\color{pygComment},
    stringstyle=\color{pygString},
    emph={heuristics_v1,hard_instance_generator_v1,heuristics},
    emphstyle=\color{pygFunction},
    showstringspaces=false,
    breaklines=true,
    columns=fullflexible,
    keepspaces=true,
    aboveskip=0pt,
    belowskip=0pt
}
\begin{tcolorbox}[colframe=black, colback=lavendermist,
boxrule=0.5pt, arc=5pt, top=0.8mm, bottom=0.8mm]
\begin{lstlisting}[style=mypython]
#TSP_GLS
def heuristics_v1(distance_matrix: np.ndarray
) -> np.ndarray:
    # Long-distance edges are typically 
    # avoided in optimal TSP_GLS solutions.

    return distance_matrix

#TSP_ACO
def heuristics_v1(distance_matrix: np.ndarray
) -> np.ndarray:
    return 1 / distance_matrix
\end{lstlisting}
\end{tcolorbox}
\vspace{-0.5em}
\begin{center}
Prompt 9: Base\_function (heuristics).
\end{center}
\vspace{-0.5em}
\begin{tcolorbox}[colframe=black, colback=lavendermist, boxrule=0.5pt, arc=5pt, top=0.8mm, bottom=0.8mm]
\begin{lstlisting}[style=mypython]
def hard_instance_generator_v1(
    distance_matrix: np.ndarray
    ) -> np.ndarray:

[task_constraint]

# The function must return all datasets directly 
# as separate numpy arrays 
# (not in a list or dictionary), 
# following exactly this format:

# Generate each dataset here...

return hard_instance_20, hard_instance_50, 
hard_instance_100, hard_instance_200, hard_instance_300,
hard_instance_400, hard_instance_500, hard_instance_600,
hard_instance_700, hard_instance_800, hard_instance_900, 
hard_instance_1000

# Do not include any test code or example calls. 
# Only output the function definition.
\end{lstlisting}
\end{tcolorbox}
\vspace{-0.5em}
\begin{center}
Prompt 10: Base\_function (hard\_instance\_generator).
\end{center}
\vspace{-0.5em}
\begin{tcolorbox}[
    colframe=black,
    colback=lavendermist,
    boxrule=0.5pt,
    arc=5pt
]

Write a hard\_instance\_generator function that generates complex and challenging [task] instances entirely in the form of adjacency matrices, to improve the performance of [task] algorithms.

The hard\_instance\_generator function must return multiple numpy arrays with the following exact shapes: 

(64, 20, 20)

(64, 50, 50)

(64, 100, 100)

(64, 200, 200)

(64, 300, 300)

(64, 400, 400)

(64, 500, 500)

(64, 600, 600)

(64, 700, 700)

(64, 800, 800)

(64, 900, 900)

(64, 1000, 1000)

Each instance must be an "adjacency matrix" representing a [task] problem with Euclidean distances or generated structural complexity (e.g., fractal, clustered, or spiral distributions). "Do not return coordinates; only return the adjacency matrices".
\end{tcolorbox}

\begin{center}
Prompt 11: Task\_constraint.
\end{center}
\begin{tcolorbox}[colframe=black, colback=lavendermist, boxrule=0.5pt, arc=5pt, top=0.8mm, bottom=0.8mm]
\begin{lstlisting}[style=mypython]
def heuristics(distance_matrix: np.ndarray
) -> np.ndarray:
\end{lstlisting}
\end{tcolorbox}

\begin{center}
Prompt 12: Function signature.
\end{center}
\begin{tcolorbox}[colframe=black, colback=lavendermist,  boxrule=0.5pt, arc=5pt]

Generate increasingly numerous and complex hard instances that specifically target the weaknesses of [task], and improve the corresponding code accordingly.

\end{tcolorbox}

\begin{center}
Prompt 13: Diagnostic anchor.
\end{center}

\lstdefinestyle{pythonstyle}{
    language=Python,
    basicstyle=\ttfamily\scriptsize,
    backgroundcolor=\color{lavendermist},
    keywordstyle=\color{blue},
    commentstyle=\color{gray},
    showstringspaces=false,
    breaklines=true,
    frame=none,
    numbers=left,
    numberstyle=\tiny,
    numbersep=5pt
}

\section{Extended Applications and Generalization Scope} 
\label{appendix:extended_applications}
The proposed \texttt{ACEvo} framework is generally applicable to optimization settings in which both problem instances and solvers can be represented as executable symbolic programs, and where relative solver performance provides a meaningful behavioral signal for guiding evolutionary feedback. Under this formulation, the adversarial co-evolution mechanism in \texttt{ACEvo} operates over a broad class of program- or string-based optimization settings in which candidate solutions are evaluated through performance-based criteria.

This formulation is consistent with prior observations on reflective evolutionary search (e.g., ReEvo~\cite{YeReEvo24}), which suggest that performance-driven symbolic evolution can be effective across diverse optimization settings without requiring access to internal solver representations. From a modeling standpoint, \texttt{ACEvo} treats solvers and instance generators uniformly as executable artifacts, enabling the co-evolutionary process to be driven solely by observed behavioral outcomes.

This positions \texttt{ACEvo} as a general mechanism for jointly amplifying instance difficulty and solver robustness under distributional shift, rather than as a method specialized for a particular task family.

\section{Experimental Details}
\label{app:exp_details}
\textbf{Computational resources.} Our framework performs co-evolutionary updates on both heuristics and problem instances, leveraging parallel execution whenever computational resources are sufficient. The total wall-clock time for each iteration of \texttt{ACEvo} is influenced by both the latency of model queries and the cost of instance evaluation. Claude 3.7 Sonnet was used as the language model backend, accessed exclusively through the Anthropic API. Across all Claude 3.7 Sonnet queries in \texttt{ACEvo}, the estimated average API cost was approximately \$0.05 per call. All evaluations were conducted on a server powered by an AMD EPYC 7742 CPU.

\begin{table}[htbp]
\centering
\caption{TSP\_GLS settings across problem sizes.}
\label{tab:gls_params}
\begin{tabular}{@{}ccc@{}}
\toprule
\textbf{Problem Size ($n$)} & \textbf{Perturbation Moves} & \textbf{Iteration Limit} \\
\midrule
20   & 5   & 73   \\
50   & 30  & 175  \\
100  & 40  & 800  \\
200  & 40  & 800  \\
300  & 40  & 800  \\
400  & 40  & 800  \\
500  & 40  & 800  \\
600  & 40  & 800  \\
700  & 40  & 800  \\
800  & 40  & 800  \\
900  & 40  & 800  \\
1000 & 40  & 800  \\
\bottomrule
\end{tabular}
\end{table}
\FloatBarrier

\subsection{Guided Local Search for the Traveling Salesman Problem}
\label{app:gls_tsp}
To ensure consistent solver behavior across varying TSP instance sizes, we set the number of perturbation moves and iteration limits as a function of the problem scale. As summarized in Table~\ref{tab:gls_params}, smaller instances are assigned fewer moves and iterations, while larger instances use fixed values to balance efficiency and effectiveness. These parameters were tuned to maintain a stable computational budget and reliable convergence behavior within our evaluation framework.

\vspace{-1em}

\subsection{Experimental Configuration}
\label{app:exp_config}
Each evolutionary run in \texttt{ACEvo} was conducted with a population
size of $10$ and $20$ iterations of adversarial co-evolution. In each
iteration, solvers and instance generators were evaluated and updated
based on performance. The system achieved stable convergence within
these iterations, demonstrating that this configuration effectively
balances efficiency and evolutionary diversity. This setup ensures
consistent evolutionary pressure while maintaining manageable
computational cost, providing a reliable foundation for all reported
results.

\subsection{Objective Value Definition in TSP\_ACO, OP\_ACO, and CVRP\_ACO}
\label{app:objective_definition}
The objective functions are determined by the underlying combinatorial optimization problems (TSP, OP, and CVRP) and are invariant to the choice of solver (see Appendix~\ref{app:ext-prelim}). 
In particular, TSP\_ACO, OP\_ACO, and CVRP\_ACO differ only in their heuristic search mechanisms, while the evaluation objective $f(\cdot)$ is consistently defined across all methods.

\section{Detailed Analysis}
\label{app:detailed_analysis}

\subsection{Detailed Description of the Visualization Protocol}
\label{appendix:vis_protocol}

\begin{figure*}[htbp]
\centering

\newcommand{\imgwidth}{0.18\textwidth}
\newcommand{\rowlabelwidth}{1.8cm}
\newcommand{\colgap}{\hspace{0.02\textwidth}}  
\newcommand{\rowgap}{1.2em}  

\makebox[\textwidth]{
\hspace*{-1.5cm}
\hspace*{\rowlabelwidth}
\begin{minipage}{\imgwidth}\centering \textbf{TSP\_GLS}\end{minipage}
\colgap
\begin{minipage}{\imgwidth}\centering \textbf{TSP\_ACO}\end{minipage}
\colgap
\begin{minipage}{\imgwidth}\centering \textbf{OP\_ACO}\end{minipage}
\colgap
\begin{minipage}{\imgwidth}\centering \textbf{CVRP\_ACO}\end{minipage}
}

\vspace{1em}

\makebox[\textwidth]{
\hspace*{-1.5cm}
\begin{minipage}[c]{\rowlabelwidth}
\centering \textbf{t-SNE}
\end{minipage}
\begin{minipage}[c]{\imgwidth}
\centering
\includegraphics[width=\linewidth]{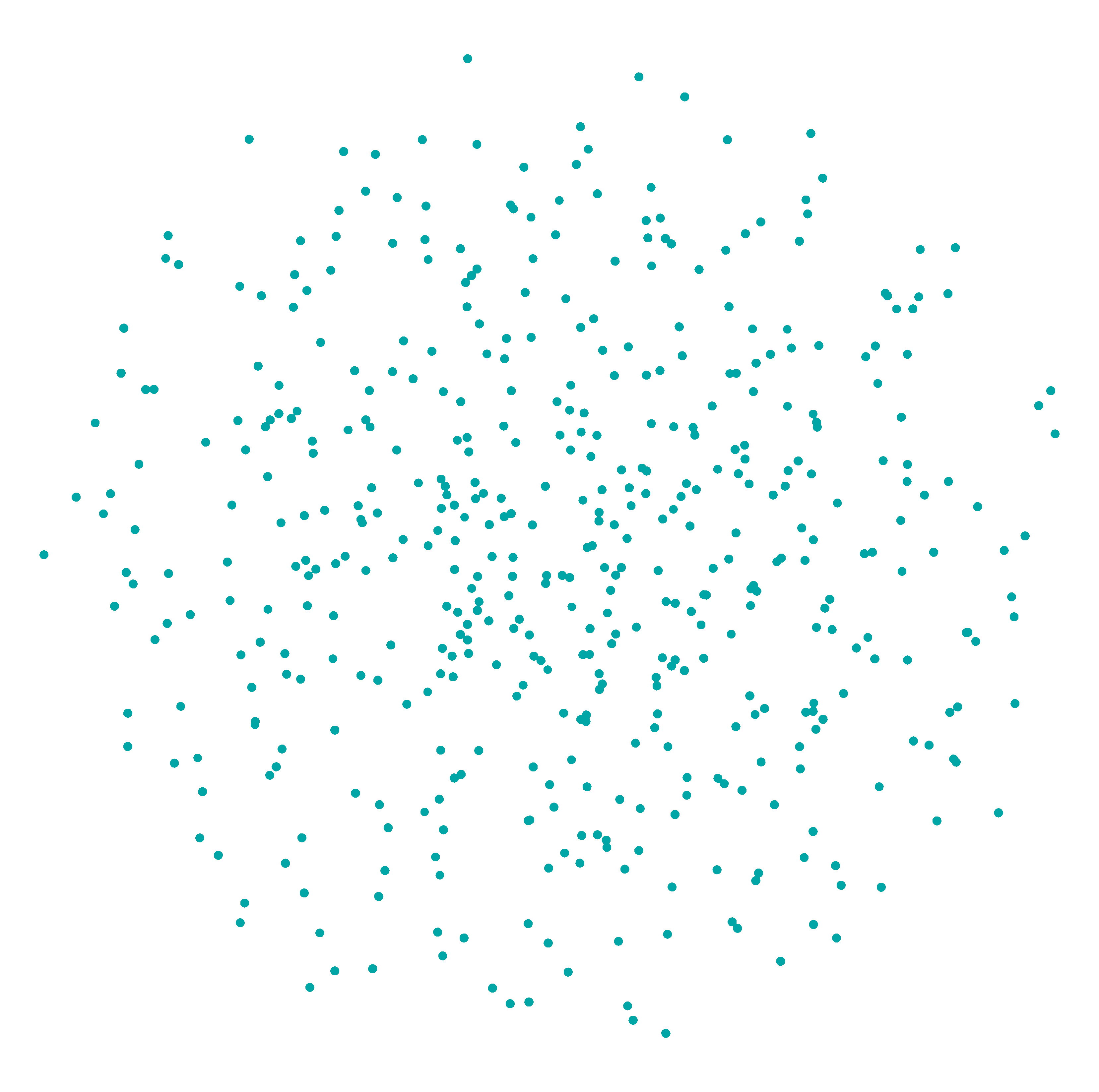}
\end{minipage}
\colgap
\begin{minipage}[c]{\imgwidth}
\centering
\includegraphics[width=\linewidth]{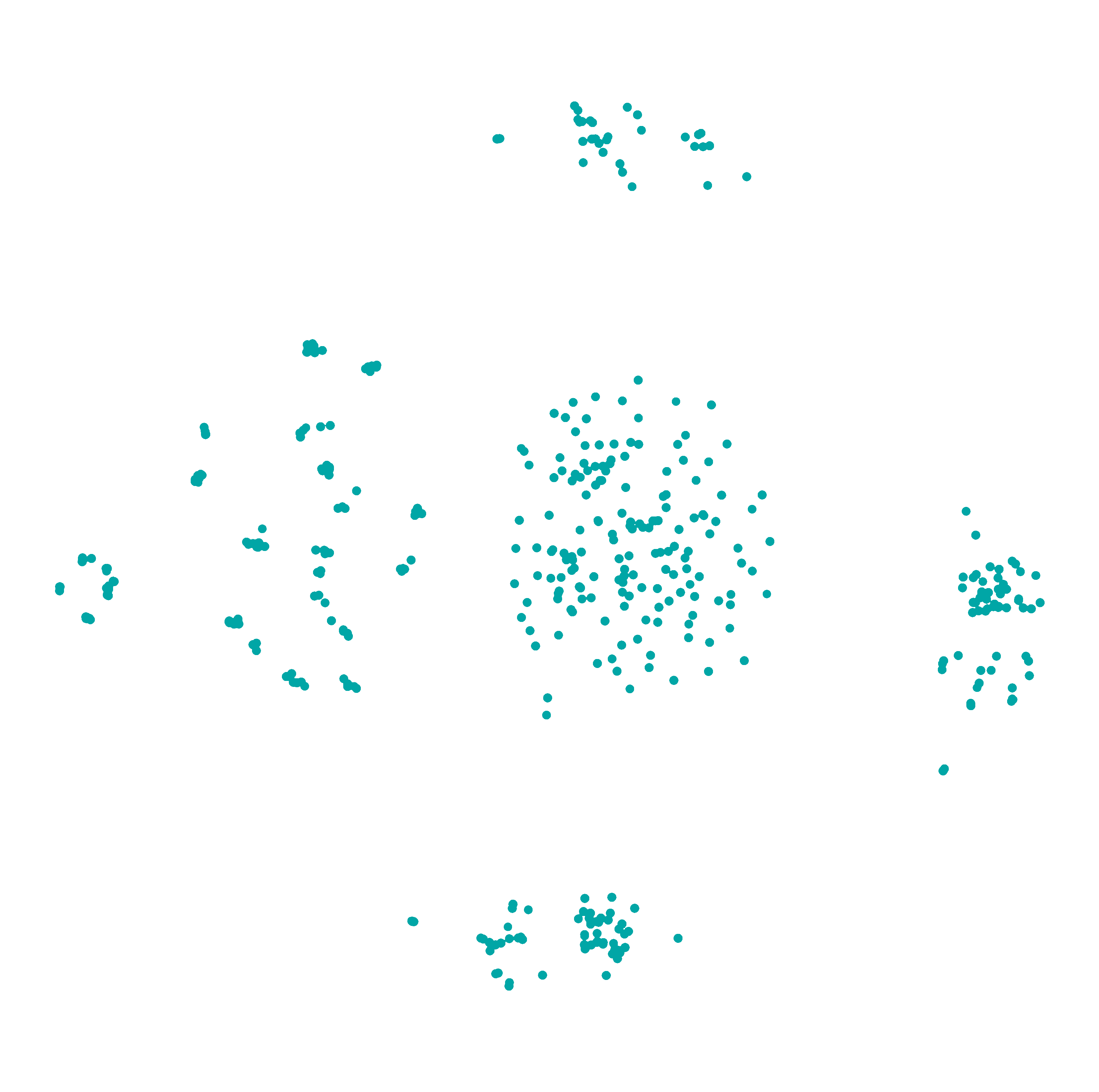}
\end{minipage}
\colgap
\begin{minipage}[c]{\imgwidth}
\centering
\includegraphics[width=\linewidth]{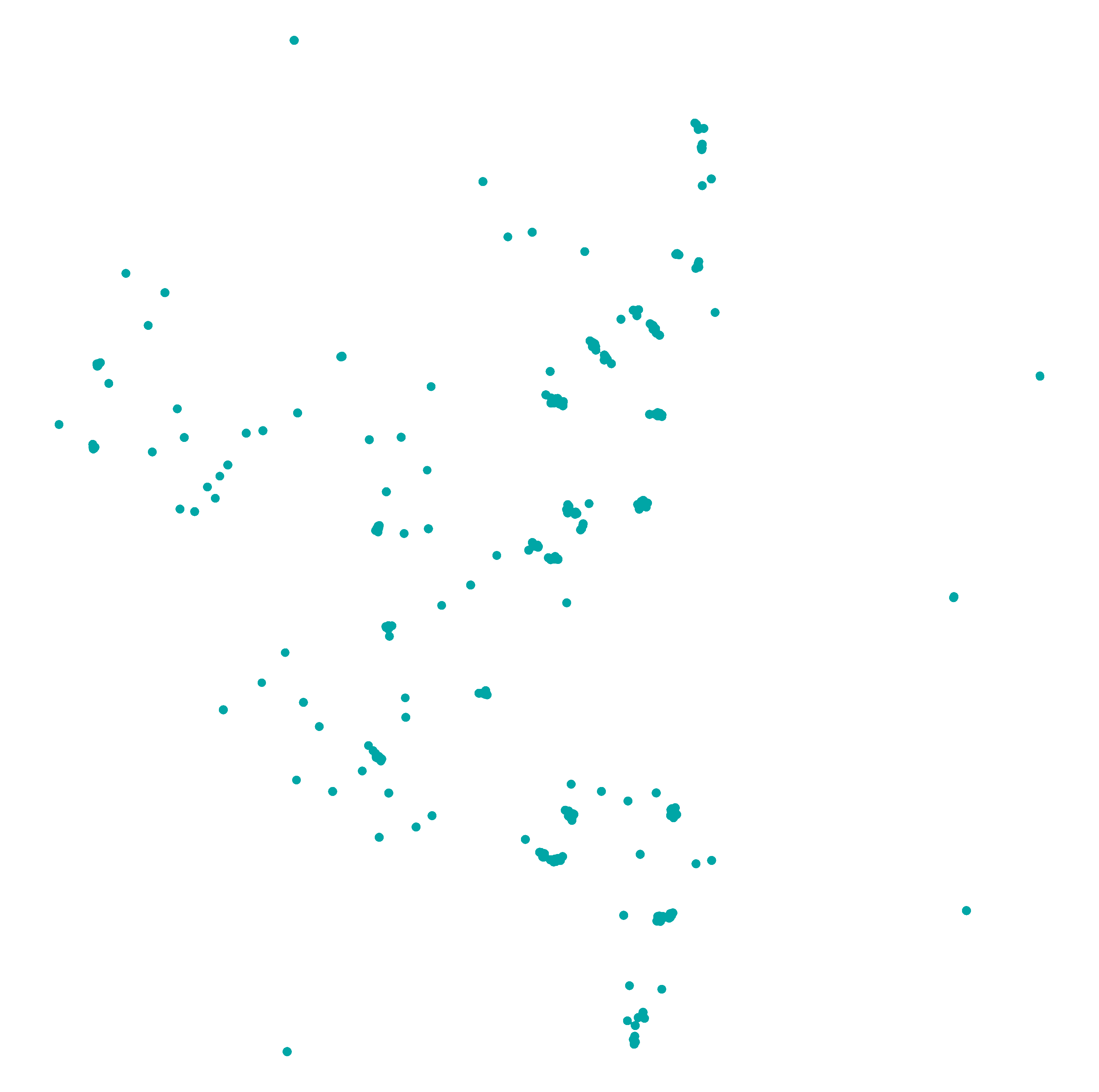}
\end{minipage}
\colgap
\begin{minipage}[c]{\imgwidth}
\centering
\includegraphics[width=\linewidth]{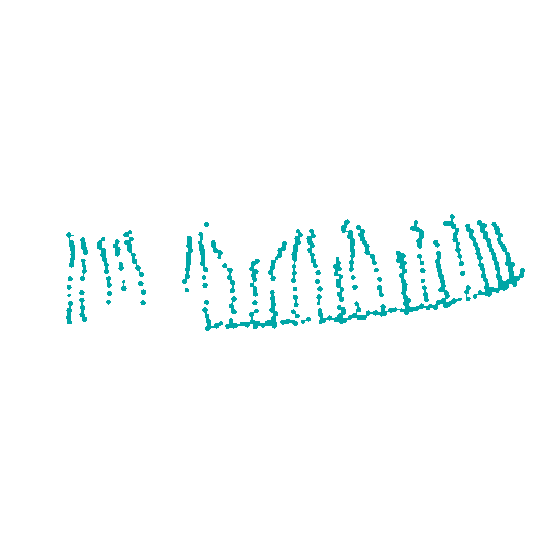}
\end{minipage}
}

\vspace{\rowgap}

\makebox[\textwidth]{
\hspace*{-1.5cm}
\begin{minipage}[c]{\rowlabelwidth}
\centering \textbf{UMAP}
\end{minipage}
\begin{minipage}[c]{\imgwidth}
\centering
\includegraphics[width=\linewidth]{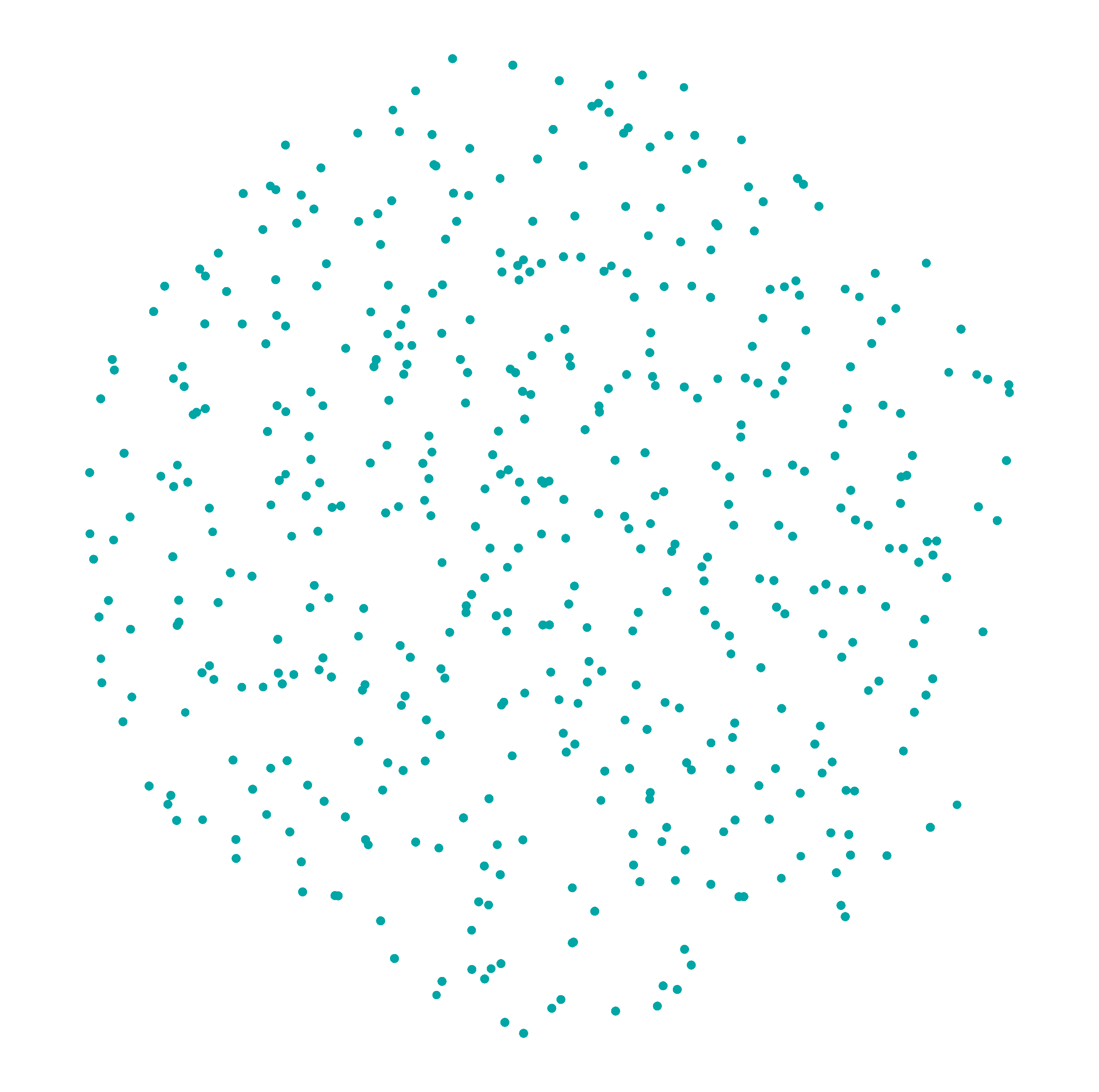}
\end{minipage}
\colgap
\begin{minipage}[c]{\imgwidth}
\centering
\includegraphics[width=\linewidth]{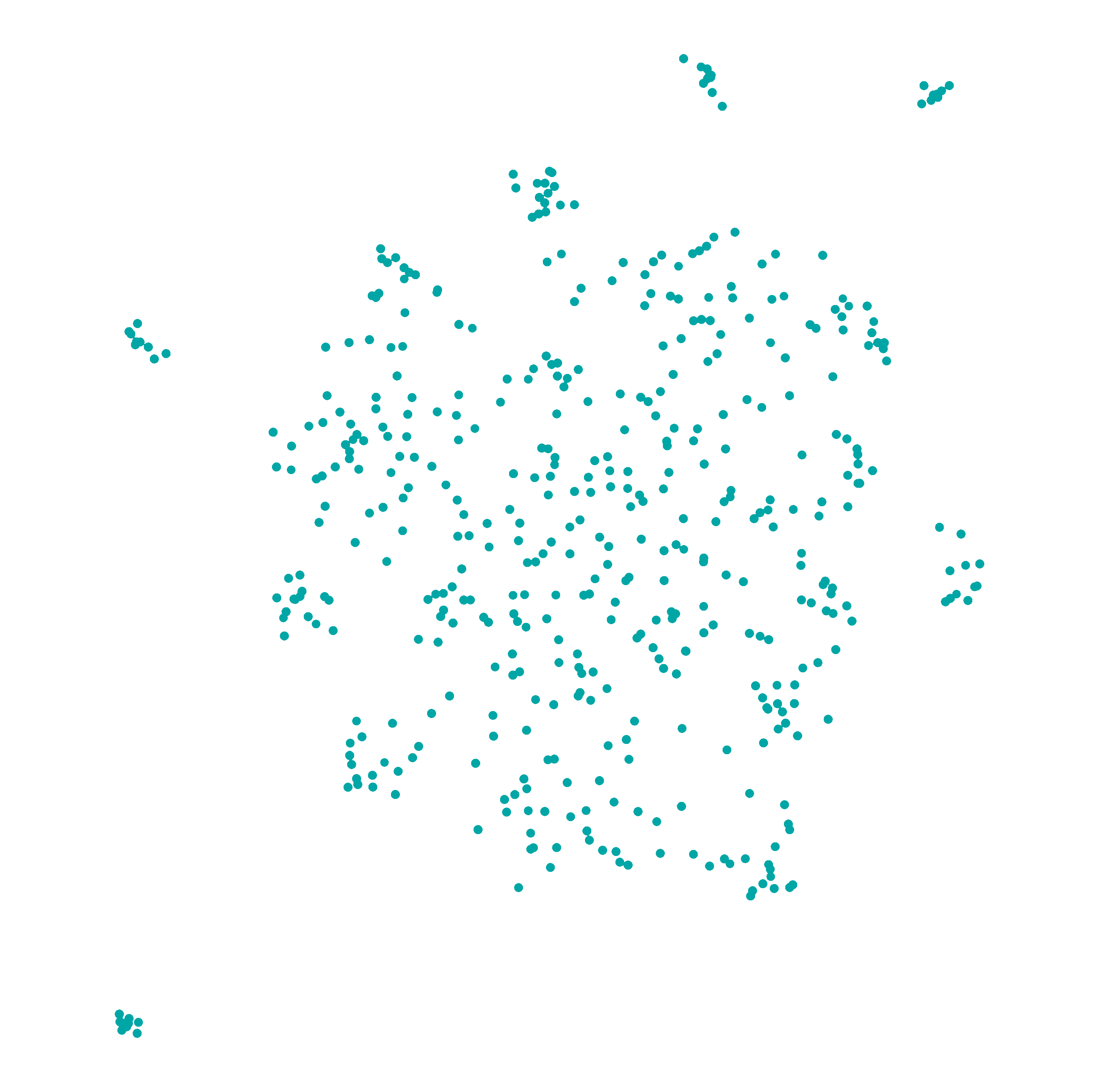}
\end{minipage}
\colgap
\begin{minipage}[c]{\imgwidth}
\centering
\includegraphics[width=\linewidth]{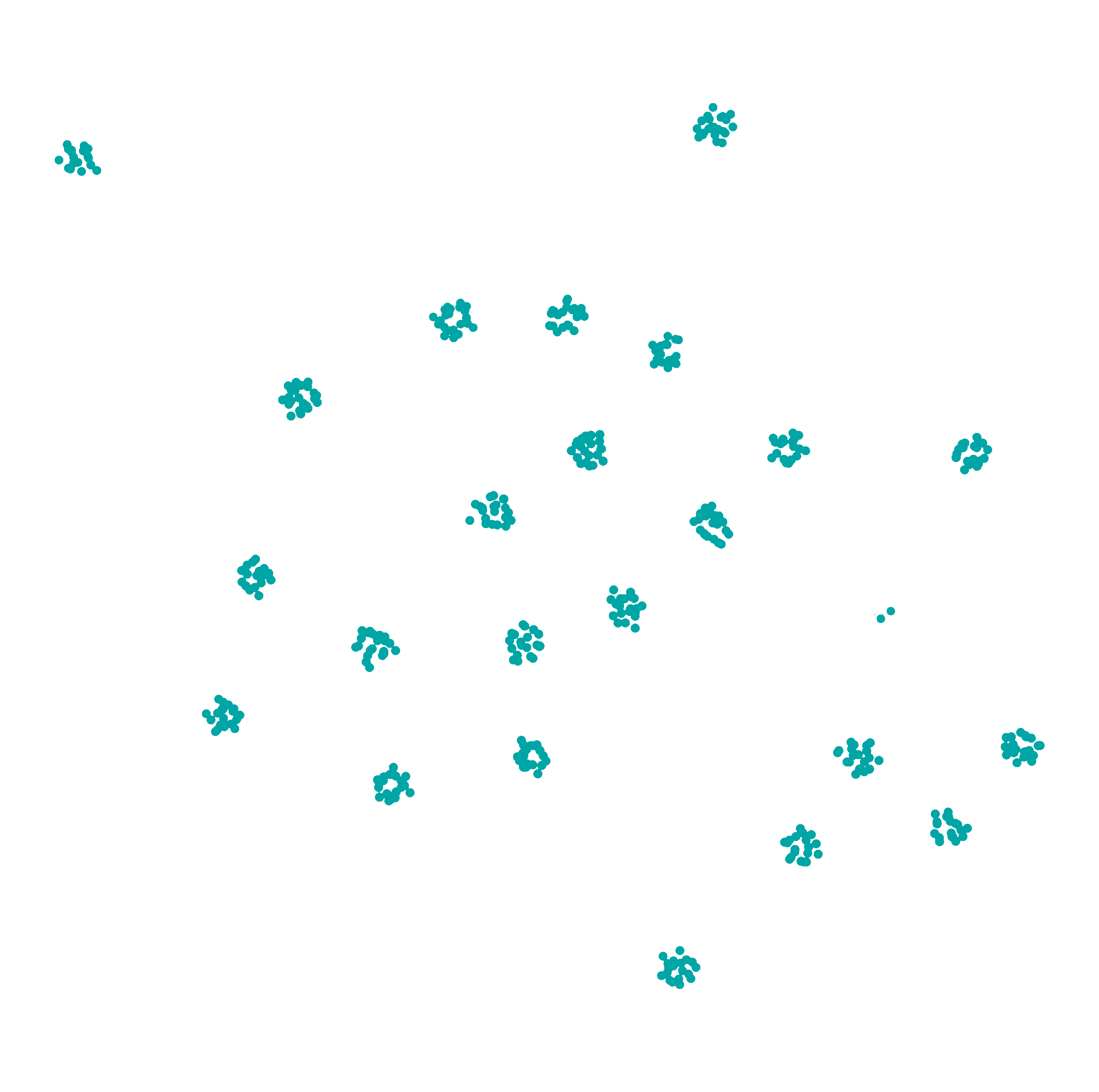}
\end{minipage}
\colgap
\begin{minipage}[c]{\imgwidth}
\centering
\includegraphics[width=\linewidth]{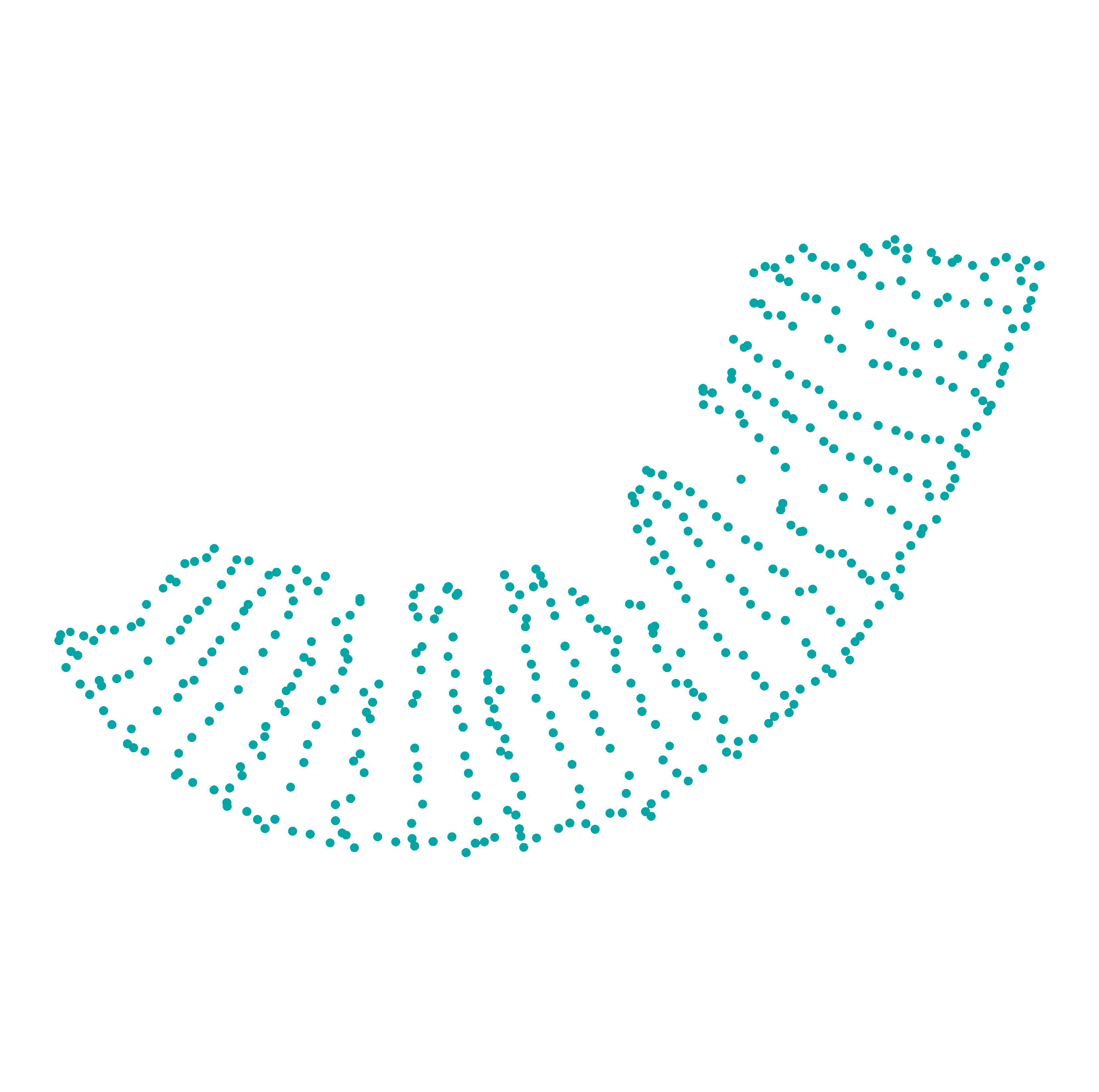}
\end{minipage}
}

\vspace{\rowgap}

\makebox[\textwidth]{
\hspace*{-1.5cm}
\begin{minipage}[c]{\rowlabelwidth}
\centering \textbf{Std}
\end{minipage}
\begin{minipage}[c]{\imgwidth}
\centering
\includegraphics[width=\linewidth]{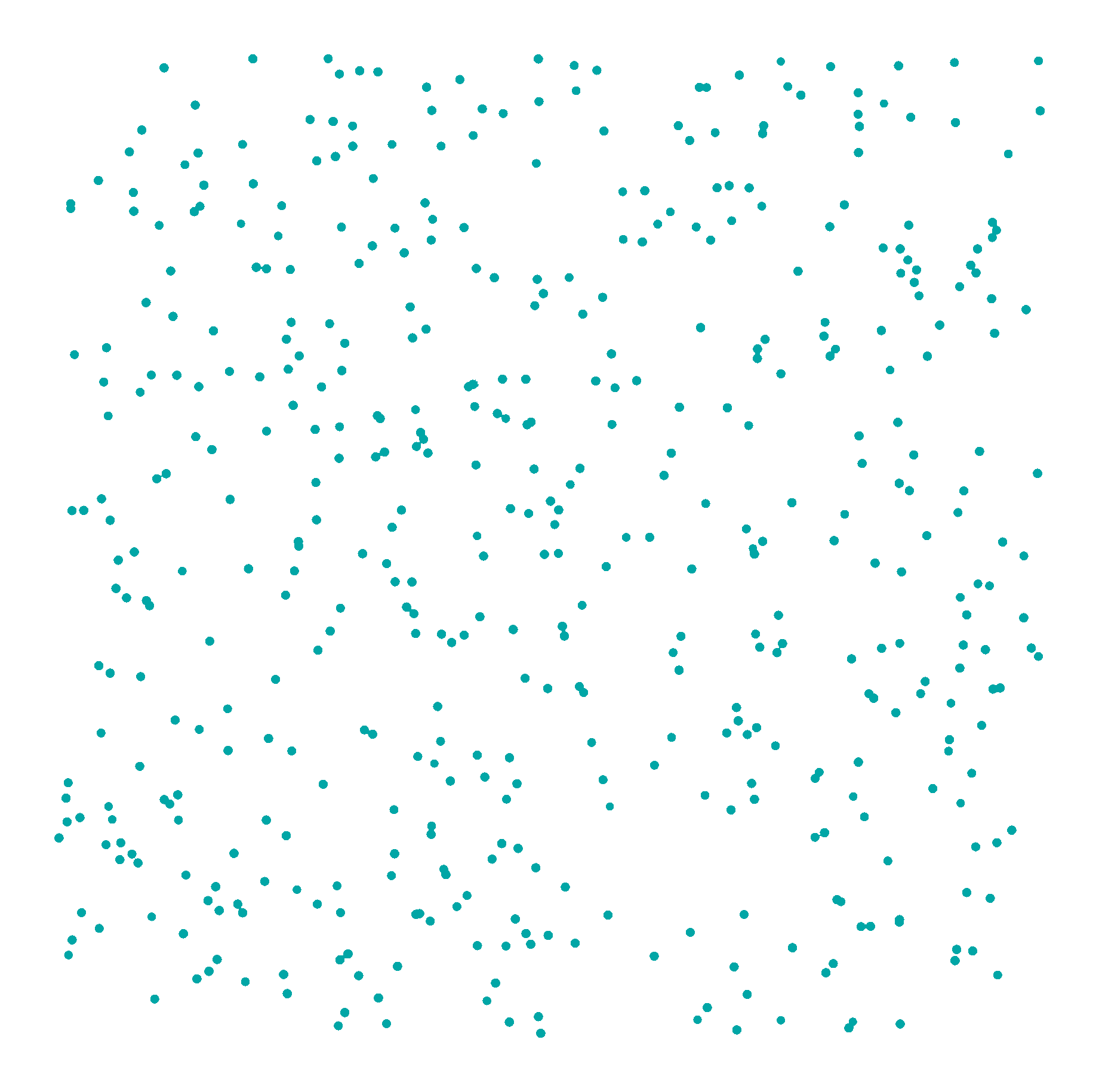}
\end{minipage}
\colgap
\begin{minipage}[c]{\imgwidth}
\centering
\includegraphics[width=\linewidth]{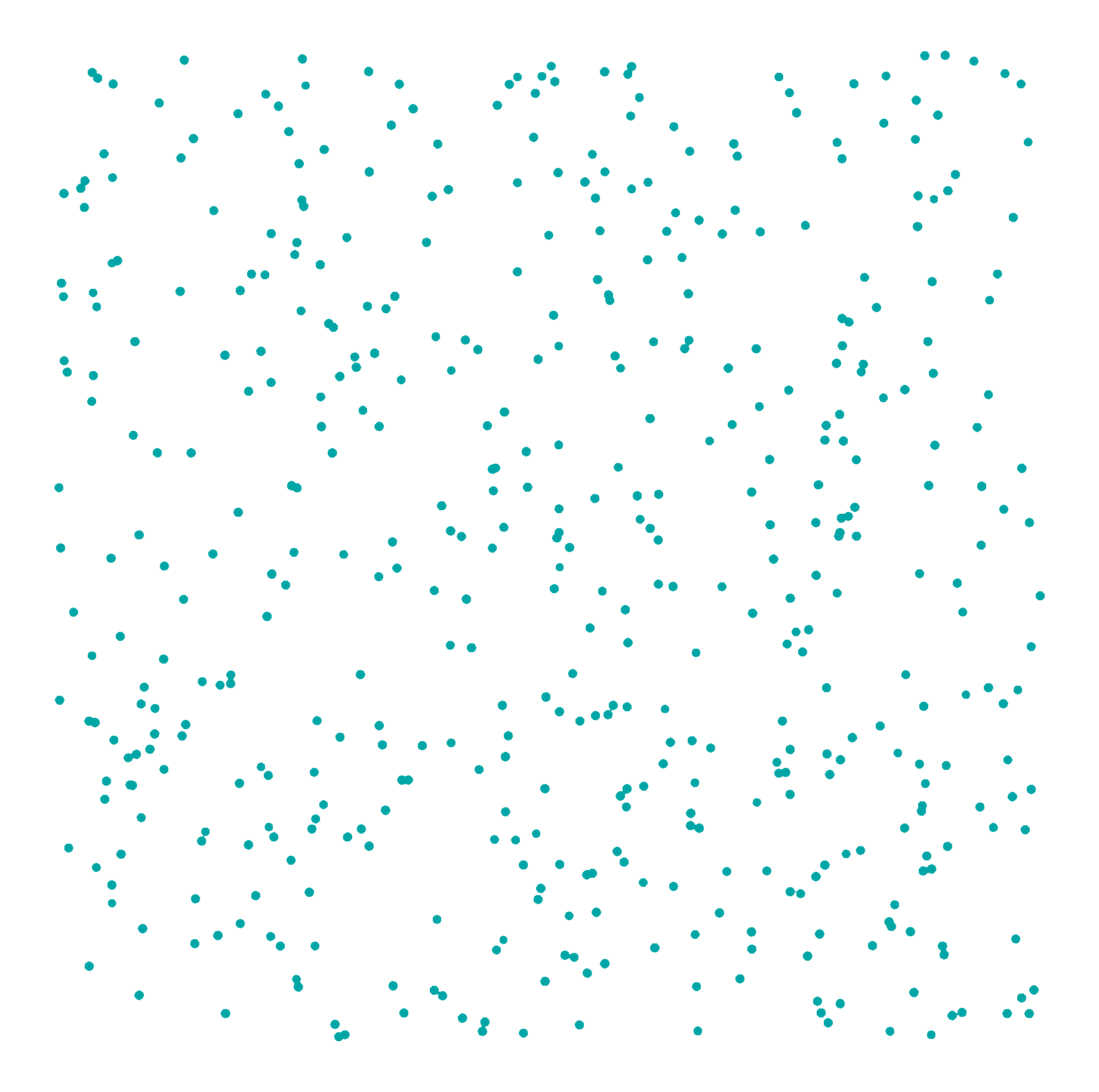}
\end{minipage}
\colgap
\begin{minipage}[c]{\imgwidth}
\centering
\includegraphics[width=\linewidth]{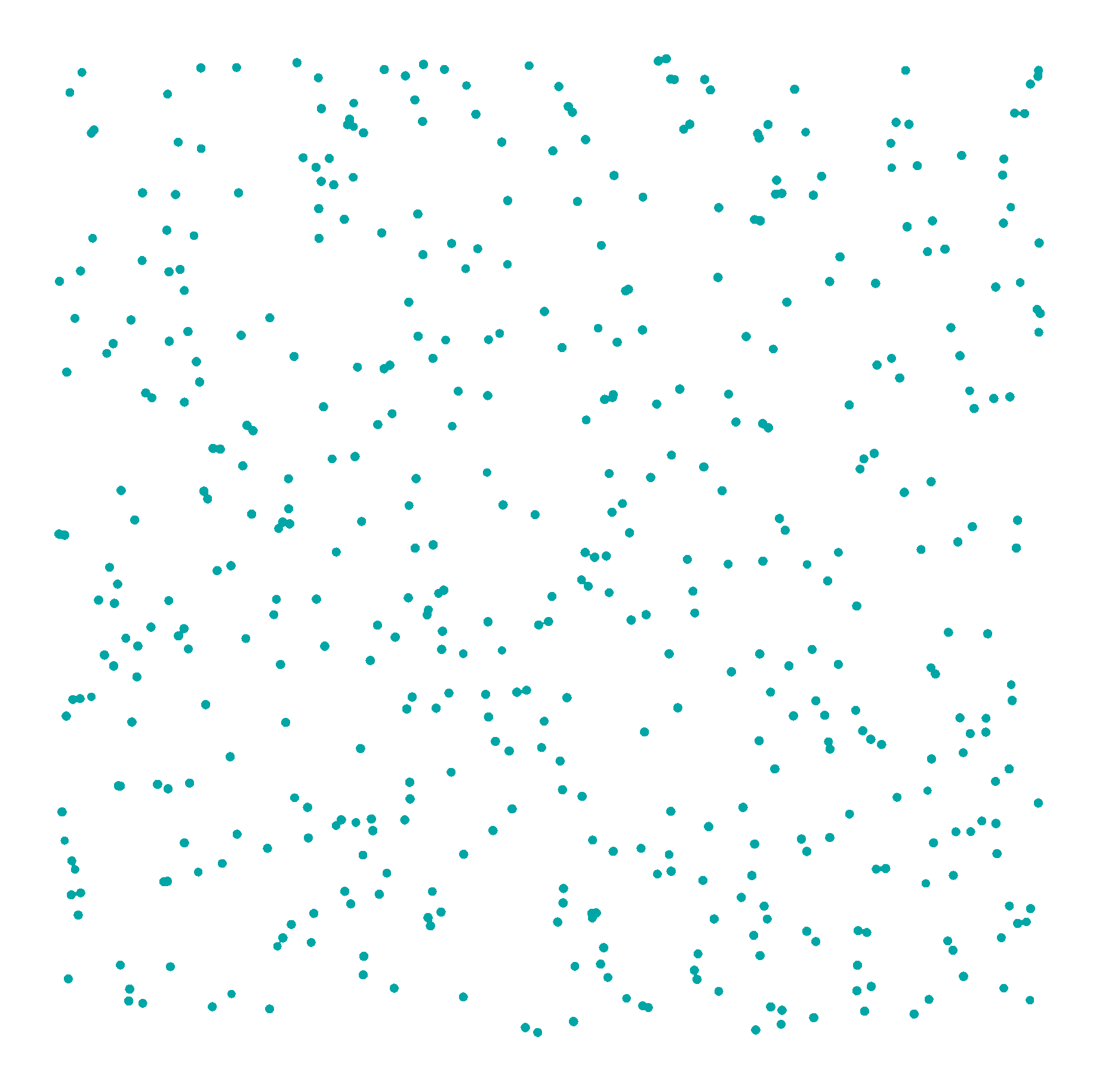}
\end{minipage}
\colgap
\begin{minipage}[c]{\imgwidth}
\centering
\includegraphics[width=\linewidth]{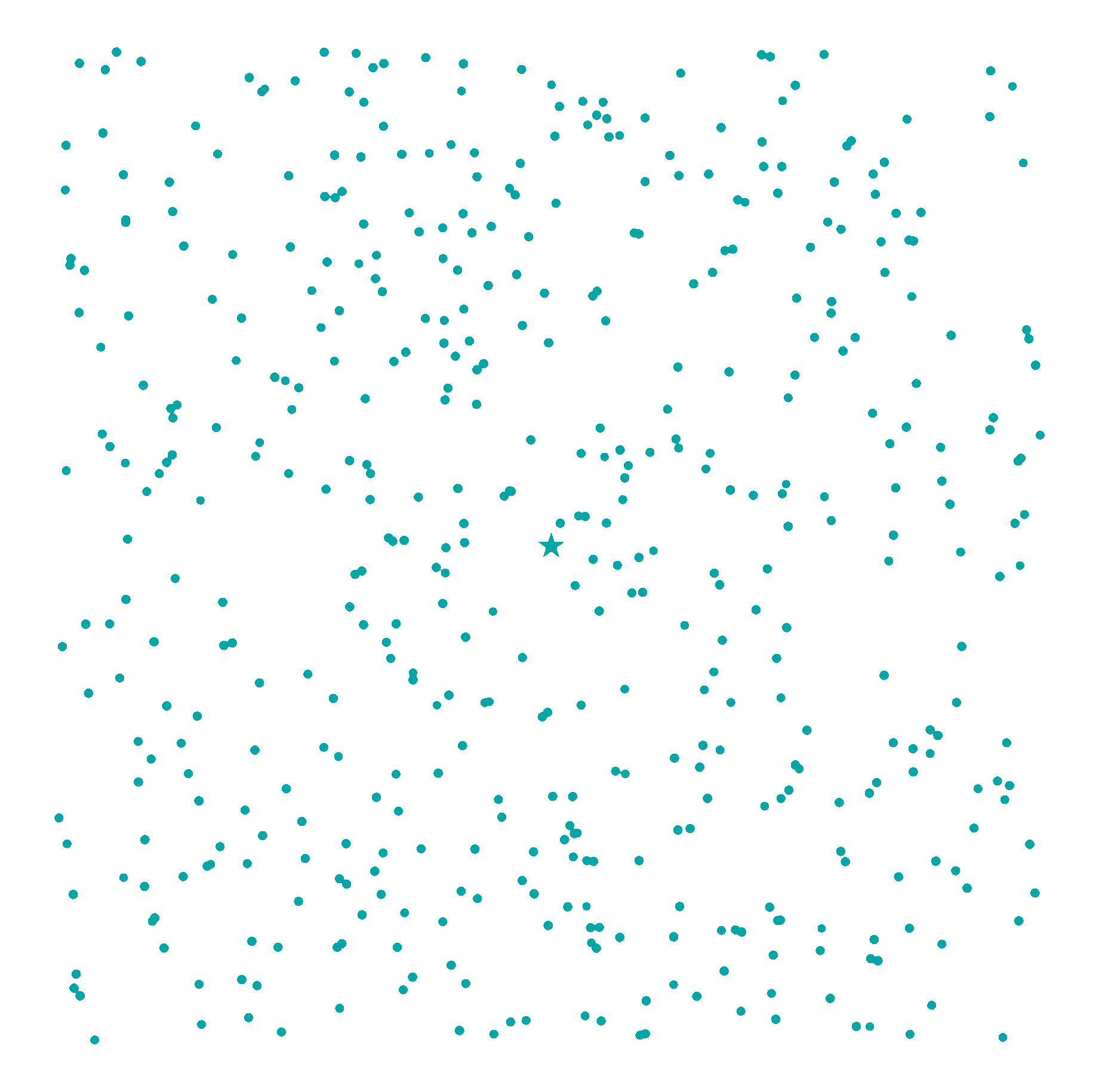}
\end{minipage}
}

\vspace{1em}

\caption{
Visualization of \texttt{ACEvo}-generated instance distributions across TSP\_GLS, TSP\_ACO, OP\_ACO, and CVRP\_ACO using t-SNE, UMAP, and standard instance layouts.
Top row shows optimization tasks; left column shows visualization methods.
The standard instance layout consists of problem instances sampled from isotropic Gaussian distributions, without any adversarial modification or structural complexity enhancement.
}
\label{fig:instance_vis_grid}
\end{figure*}

Figure~\ref{fig:instance_vis_grid} provides an overview of the visualization grid used to analyze instance structures. The first, second, and third rows correspond to t\textendash SNE~\cite{van2008visualizing}, UMAP~\cite{2018arXivUMAP}, and standard two\textendash dimensional layouts, respectively. The t\textendash SNE~\cite{van2008visualizing} and UMAP~\cite{2018arXivUMAP} rows illustrate the distributions of \texttt{ACEvo}\textendash generated hard instances, while the bottom (Std) row depicts standard datasets for reference. The columns represent TSP\_GLS, TSP\_ACO, OP\_ACO, and CVRP\_ACO from left to right, with CVRP\_ACO additionally incorporating capacity constraints and depot\textendash centered routing structures. Each scatter plot shows a two\textendash dimensional embedding of instance representations obtained via nonlinear manifold projection, enabling qualitative comparison between \texttt{ACEvo}\textendash generated hard instances and datasets such as ReEvo~\cite{YeReEvo24}. Compared with conventional datasets, the \texttt{ACEvo} instances exhibit denser local clusters, broader dispersion, and clearer structural boundaries, revealing adaptive dynamics of adversarial co\textendash evolution—where the generator continually diversifies its outputs as the solver improves. This emergent diversity underscores \texttt{ACEvo}'s ability to break dataset homogeneity and foster coadaptation between instance generation and solver evolution.

\FloatBarrier

\begin{figure}[htbp]
    \centering
    \includegraphics[width=0.8\textwidth]{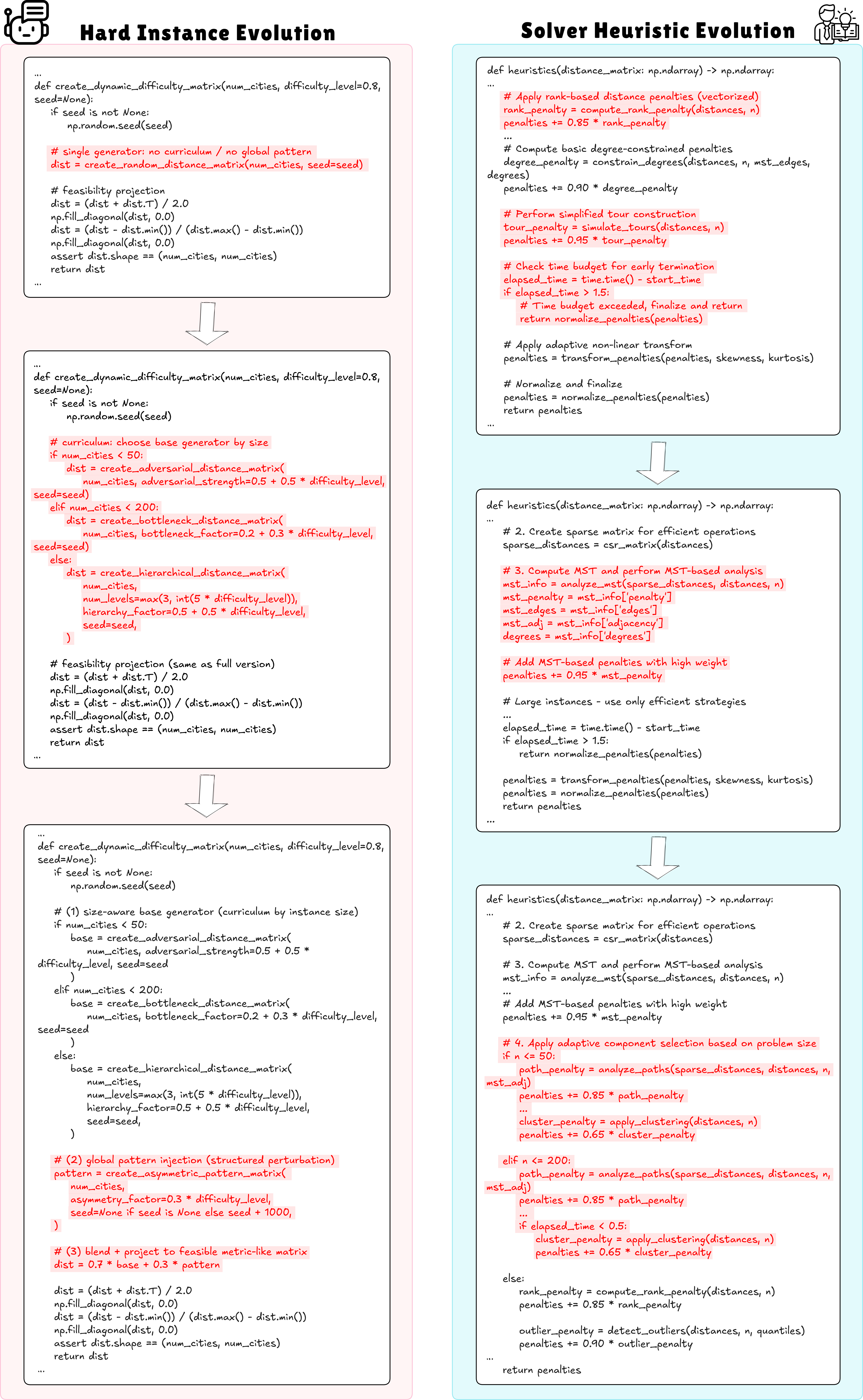}
    \caption{Representative program fragments from different stages of the ACEvo co-evolution process, illustrating hard instance evolution (left) and solver heuristic evolution (right).}

    \label{fig:ex}
\end{figure}
\FloatBarrier

\subsection{Evolution Process}
\label{subsec:evolution-process}
Figure~\ref{fig:ex} illustrates representative program fragments sampled at different stages of the ACEvo co-evolution process. 
The figure is intended to clarify how instance generators and heuristic scoring functions evolve structurally through mutation and selection, rather than to enumerate all implementation details.
Each fragment corresponds to a concrete executable variant retained after selection in a specific evolutionary stage.
Only representative snapshots are shown, as the full evolutionary trajectory consists of many intermediate mutations.
The examples are chosen to highlight structural changes that persist across multiple iterations, rather than transient or degenerate variants.

\vspace{-1em}
\paragraph{Hard Instance Evolution (left).}
The initial generator constructs distance matrices using a single randomized procedure, without explicit global structure or scale awareness. 
In later variants, the generator adopts a size-conditioned design, where different base construction mechanisms are selected according to the number of cities. 
This enables distinct structural regimes to emerge at different problem scales, such as adversarial perturbations for small instances, bottlenecked structures for medium instances, and hierarchical layouts for larger instances. 
In the final variant, an explicit global pattern component is injected and blended with the base generator, introducing structured asymmetries while preserving metric feasibility through normalization and projection. 
Across all stages, feasibility constraints are enforced to ensure valid TSP instances.
\vspace{-1em}
\paragraph{Solver Heuristic Evolution (right).}
The heuristic function evolves from simple, uniform penalty assignments toward progressively more structured analyses of the distance matrix. 
Intermediate variants introduce rank-based penalties, degree-aware signals, and approximate tour simulations to capture local and mid-range structure. 
Later versions incorporate sparse representations and minimum spanning tree (MST)--based analysis to extract global connectivity information efficiently. 
In the final stage, the heuristic adaptively activates different analytical components---such as path-based penalties, clustering signals, or outlier detection---based on problem size and time constraints, allowing scale-dependent trade-offs between expressiveness and computational cost.

\subsection{Statistical Robustness and Sensitivity Analysis}
\label{app:robustness}

To assess the robustness of \texttt{ACEvo} under stochastic generation,
we perform independent runs with different random seeds for the
ablation experiments reported in Table~\ref{tab:ablation}.
No new experimental settings are introduced; each run follows the same
experimental protocol and differs only in random initialization and
stochastic sampling. Each run involves independent sampling from the large language model,
together with randomized mutation, reflection, and selection operations
during the co-evolutionary process.
This setup provides an empirical sensitivity analysis with respect to
random seeds and stochastic LLM sampling.
Table~\ref{tab:ablation} reports the mean performance and one standard
deviation across these runs.
This repeated evaluation allows us to quantify the extent to which the observed performance depends on stochastic fluctuations rather than systematic properties of the proposed framework.

The full \texttt{ACEvo} configuration exhibits relatively small
run-to-run variation in both hard-instance generation and solver
performance.
The corresponding standard deviations remain lower than those observed
for most ablated variants, indicating that the complete framework
maintains stable performance across different stochastic realizations.
In contrast, removing adversarial feedback, reflection, or mutation
generally increases performance variability, in addition to degrading
the mean results.
This pattern further indicates that the individual components contribute not only to higher average performance but also to reducing the sensitivity of the co-evolutionary process to random initialization and sampling noise.

These findings suggest that the gains of \texttt{ACEvo} are not driven
by a single favorable random seed or sampling trajectory.
Rather, the combination of adversarial feedback, evolutionary
reflection, and mutation produces consistently strong outcomes across
independent runs, demonstrating robustness to the inherent stochasticity
of LLM-based generation and evolutionary search.
The consistency of these results across independent repetitions therefore strengthens the statistical reliability of the conclusions drawn from the main ablation study.

\begin{table}[htbp]
\centering
\caption{OP\_ACO objective values under different heuristic--instance combinations. Higher is better.}
\label{tab:op_aco_obj}

\setlength{\tabcolsep}{4pt}
\renewcommand{\arraystretch}{1.15}
\scriptsize

\resizebox{\linewidth}{!}{%
\begin{tabular}{@{}c c c@{}}
\toprule
\textbf{Problem Size ($n$)} &
\makecell{\textbf{heuristics (ReEvo)}\cite{YeReEvo24}\\$\oplus$ \textbf{hard instance (ACEvo)}} &
\makecell{\textbf{heuristics (ACEvo)}\\$\oplus$ \textbf{hard instance (ACEvo)}} \\
\midrule
100  & 7.883 & \textbf{7.923}  \\
200  & 13.862 & \textbf{14.157} \\
300  & 17.461 & \textbf{17.635} \\
400  & 17.773 & \textbf{18.662} \\
500  & 24.944 & \textbf{25.388} \\
600  & 22.655 & \textbf{23.532} \\
700  & 16.457 & \textbf{17.516} \\
800  & 23.210 & \textbf{24.752} \\
900  & 20.217 & \textbf{20.760} \\
1000 & 20.061 & \textbf{21.205} \\
\bottomrule
\end{tabular}%
}
\end{table}
\FloatBarrier

\begin{table}[htbp]
\centering
\caption{CVRP\_ACO objective values under different heuristic--instance combinations. Lower is better.}
\label{tab:cvrp_aco_obj}

\setlength{\tabcolsep}{3pt}
\renewcommand{\arraystretch}{1.05}
\scriptsize

\resizebox{\linewidth}{!}{%
\begin{tabular}{@{}c c c@{}}
\toprule
\textbf{Problem Size ($n$)} &
\makecell{\textbf{heuristics (ReEvo)}\cite{YeReEvo24}\\$\oplus$ \textbf{hard instance (ACEvo)}} &
\makecell{\textbf{heuristics (ACEvo)}\\$\oplus$ \textbf{hard instance (ACEvo)}} \\
\midrule
100  & 15.878 & \textbf{15.616}  \\
200  & 30.288 & \textbf{29.853} \\
300  & 44.153 & \textbf{43.915} \\
400  & 57.984 & \textbf{57.128} \\
500  & 71.895 & \textbf{70.932} \\
600  & 85.552 & \textbf{84.695} \\
700  & 99.122 & \textbf{97.920} \\
800  & 112.340 & \textbf{110.835} \\
900  & 126.612 & \textbf{125.317} \\
1000 & 139.767 & \textbf{138.566} \\
\bottomrule
\end{tabular}%
}
\vspace{-0.5em}
\end{table}
\FloatBarrier

\vspace{-0.5em}

\subsection{Solver--instance synergy on the Orienteering Problem.}
\vspace{-0.5em}

\label{appendix:op}

This appendix provides supplementary experimental results for solver--instance synergy on the Orienteering Problem. In particular, Table~\ref{tab:op_aco_obj} reports the performance of different heuristic--instance combinations for OP\_ACO. These results complement the discussion in the main text and further support the effectiveness of \texttt{ACEvo} in jointly evolving heuristics and hard instance distributions.

\vspace{-0.5em}

\subsection{Solver--instance synergy on the Capacitated Vehicle Routing Problem.}
\vspace{-0.5em}

\label{appendix:cvrp}
Table~\ref{tab:cvrp_aco_obj} summarizes the objective values obtained by different heuristic--instance combinations on the Capacitated Vehicle Routing Problem solved via Ant Colony Optimization (CVRP\_ACO). Across all evaluated problem sizes, heuristics co-evolved with \texttt{ACEvo} consistently yield lower routing costs on the same hard instances than heuristics generated by ReEvo~\cite{YeReEvo24}, indicating a systematic performance advantage under capacity-constrained routing scenarios.

This advantage is particularly notable given the additional feasibility constraints imposed by CVRP, where demand satisfaction and vehicle capacity jointly shape route construction and cost accumulation. The results suggest that solvers evolved within the \texttt{ACEvo} framework better internalize these structural constraints under jointly constrained and highly nonconvex settings when operating on adversarial instance distributions. Together, these findings provide complementary evidence that the proposed co-evolutionary paradigm generalizes beyond unconstrained or reward-based routing problems, extending effectively to capacity-constrained combinatorial optimization tasks.

Beyond absolute performance differences, the relative ordering between heuristics remains stable as problem size increases. As the number of customers grows, both methods exhibit higher objective values due to increased routing complexity; however, the performance gap in favor of \texttt{ACEvo}-evolved heuristics persists across all evaluated scales. This consistent trend suggests that the observed solver--instance synergy is not driven by isolated problem sizes or favorable configurations, but reflects a systematic advantage that scales with the complexity of capacity-constrained routing tasks.

\begin{table}[htbp]
\centering
\caption{Ablation study on the evolutionary components of \texttt{ACEvo}. 
We report the mean performance gap (\%) $\pm$ one standard deviation across 10 independent runs, 
where each run averages the gap over all test instances. 
Lower is better.}
\label{tab:ablation}
\renewcommand{\arraystretch}{1.05}
\setlength{\tabcolsep}{4.2pt}
\small
\resizebox{\linewidth}{!}{%
\begin{tabular}{lcc}
\toprule
\textbf{Method} & \textbf{Hard Instance Gap (\%)} & \textbf{Solver Gap (\%)} \\
\midrule
Full \texttt{ACEvo} (all enabled)  & \textbf{10.43 $\pm$ 0.19} & \textbf{8.10 $\pm$ 0.16} \\
w/o Adversarial Feedback          & 7.76 $\pm$ 0.44           & 11.43 $\pm$ 0.57        \\
w/o Reflection                    & 7.74 $\pm$ 0.26           & 10.21 $\pm$ 0.34        \\
w/o Mutation                      & 8.62 $\pm$ 0.31           & 9.33 $\pm$ 0.23         \\
\bottomrule
\end{tabular}
}
\end{table}
\FloatBarrier

\subsection{Ablation Study on Evolutionary Components}
\label{app:ablation_components}

Table~\ref{tab:ablation} presents an ablation study on the key evolutionary components of \texttt{ACEvo}, including adversarial feedback, reflection, and mutation. The results are reported as mean gap values with standard deviations over 10 independent runs. Overall, the full \texttt{ACEvo} framework achieves the strongest hard-instance generation performance while also maintaining the best solver-side performance, indicating that these components contribute jointly to the effectiveness and stability of the co-evolutionary process.

\subsection{Adversarial Co-Evolution Procedure of \texttt{ACEvo}}

\label{app:acevo-pseudocode}

Algorithm~\ref{alg:acevo} summarizes the overall adversarial co-evolutionary procedure of \texttt{ACEvo}, as illustrated in Figure~\ref{fig:framework}. The framework maintains two interacting populations: a population of heuristic solvers and a population of hard-instance generators. At each iteration, the current generators produce candidate instance batches on which the solver population is evaluated, thereby identifying both strong solvers and generators that induce substantial difficulty. Conditioned on the hardest instances, the solver branch mutates existing heuristics, evaluates their behavioral improvements, and selects candidates that preserve both effectiveness and diversity. Symmetrically, conditioned on the current solver population, the instance branch mutates generator programs to construct increasingly challenging instances that expose solver vulnerabilities more sharply.

To guide the evolutionary updates, the framework extracts structured feedback from the observed behavioral differences between parent and mutated programs. This feedback is incorporated into subsequent optimization steps, enabling both populations to progressively adapt toward more effective search strategies and more challenging instance distributions. By repeatedly alternating mutation, selection, and feedback-driven adaptation on both sides, \texttt{ACEvo} induces a closed-loop adversarial dynamic in which solver robustness and instance hardness are progressively strengthened together. The procedure ultimately returns the strongest evolved solver and the hardest evolved instance generator obtained during co-evolution.

\begin{algorithm}[t]
\caption{\texttt{ACEvo}: Adversarial Co-Evolution of Heuristic Solvers and Hard Instance Generators}
\label{alg:acevo}
\small

\Input{
problem specification $\mathcal{T}$; heuristic language model $L_{\mathrm{heur}}$; instance language model $L_{\mathrm{inst}}$; 
initial prompts $P_{\mathrm{heur}}^{(0)}$ and $P_{\mathrm{inst}}^{(0)}$; 
heuristic population size $M_h$; generator population size $M_g$; 
evaluation budget $B$; number of iterations $T$
}
\Output{best heuristic solver $h^\star$ and best hard-instance generator $g^\star$}

\Init{
sample initial heuristic population 
$\mathcal{H}^{(0)}=\{h_1^{(0)},\dots,h_{M_h}^{(0)}\}\sim L_{\mathrm{heur}}(P_{\mathrm{heur}}^{(0)})$\;
sample initial generator population 
$\mathcal{G}^{(0)}=\{g_1^{(0)},\dots,g_{M_g}^{(0)}\}\sim L_{\mathrm{inst}}(P_{\mathrm{inst}}^{(0)})$\;
initialize reflection memories $\mathcal{R}_h^{(0)}\gets\emptyset$ and $\mathcal{R}_g^{(0)}\gets\emptyset$\;
}

\For{$t\gets1$ \KwTo $T$}{

    \tcc{Phase 1: Cross-evaluation between solver and generator populations}
    \ForEach{$g \in \mathcal{G}^{(t-1)}$}{
        sample an instance batch $\mathcal{I}_g=\{I_1,\dots,I_B\}$ using $g$\;
    }
    
    \ForEach{$h \in \mathcal{H}^{(t-1)}$}{
        evaluate $h$ on all batches $\mathcal{I}_g$ generated by $\mathcal{G}^{(t-1)}$\;
        compute performance scores $f(h(I))$ and relative hardness values\;
    }

    select the currently hardest generator $\hat{g}^{(t-1)}$ according to induced solver gap\;
    select the currently strongest solver $\hat{h}^{(t-1)}$ according to performance on hard instances\;

    \tcc{Phase 2: Heuristic evolution under adversarial hard instances}
    construct hard batch $\hat{\mathcal{I}}^{(t-1)}$ using $\hat{g}^{(t-1)}$\;
    initialize mutated heuristic set $\tilde{\mathcal{H}}^{(t)}\gets\emptyset$\;

    \ForEach{$h \in \mathcal{H}^{(t-1)}$}{
        compose heuristic mutation prompt using $\mathcal{T}$, $h$, $\hat{\mathcal{I}}^{(t-1)}$, and $\mathcal{R}_h^{(t-1)}$\;
        generate mutated solver $\tilde{h}\sim L_{\mathrm{heur}}(\tilde{P}_{\mathrm{heur}})$\;
        evaluate both $h$ and $\tilde{h}$ on $\hat{\mathcal{I}}^{(t-1)}$\;
        derive evolutionary reflection $r_h$ from their behavioral difference, failure modes, and performance traces\;
        add $\tilde{h}$ to $\tilde{\mathcal{H}}^{(t)}$\;
        update heuristic reflection memory with $r_h$\;
    }

    select high-performing and diverse solvers from 
    $\mathcal{H}^{(t-1)} \cup \tilde{\mathcal{H}}^{(t)}$
    to form $\mathcal{H}^{(t)}$\;

    \tcc{Phase 3: Hard-instance evolution against current solvers}
    choose anchor solver set $\bar{\mathcal{H}}^{(t)} \subseteq \mathcal{H}^{(t)}$\;
    initialize mutated generator set $\tilde{\mathcal{G}}^{(t)}\gets\emptyset$\;

    \ForEach{$g \in \mathcal{G}^{(t-1)}$}{
        compose generator mutation prompt using $\mathcal{T}$, $g$, $\bar{\mathcal{H}}^{(t)}$, and $\mathcal{R}_g^{(t-1)}$\;
        generate mutated generator $\tilde{g}\sim L_{\mathrm{inst}}(\tilde{P}_{\mathrm{inst}})$\;
        sample candidate hard-instance batch $\mathcal{I}_{\tilde{g}}$ using $\tilde{g}$\;
        evaluate hardness of $\mathcal{I}_{\tilde{g}}$ against $\bar{\mathcal{H}}^{(t)}$\;
        derive evolutionary reflection $r_g$ from hardness increase, structural changes, and solver failures\;
        add $\tilde{g}$ to $\tilde{\mathcal{G}}^{(t)}$\;
        update generator reflection memory with $r_g$\;
    }

    select harder and structurally diverse generators from
    $\mathcal{G}^{(t-1)} \cup \tilde{\mathcal{G}}^{(t)}$
    to form $\mathcal{G}^{(t)}$\;

    \tcc{Phase 4: Adversarial co-evolution}
    recompute cross-play between $\mathcal{H}^{(t)}$ and $\mathcal{G}^{(t)}$\;
    preserve stronger solvers and harder generators to sustain adversarial pressure\;
}
select
$h^\star = \arg\min_{h \in \mathcal{H}^{(T)}} \mathrm{HardInstanceGap}(h)$\;
select
$g^\star = \arg\max_{g \in \mathcal{G}^{(T)}} \mathrm{Hardness}(g \mid \mathcal{H}^{(T)})$\;

\Return{$h^\star, g^\star$}
\end{algorithm}

\end{document}

%% file: math_commands.tex
\usepackage{amsmath,amsfonts,bm}

\def\eqref#1{equation~\ref{#1}}

\def\1{\bm{1}}

\DeclareMathAlphabet{\mathsfit}{\encodingdefault}{\sfdefault}{m}{sl}
\SetMathAlphabet{\mathsfit}{bold}{\encodingdefault}{\sfdefault}{bx}{n}

